\documentclass[letterpaper]{article} 
\usepackage{aaai2026}  
\usepackage{times}  
\usepackage{helvet}  
\usepackage{courier}  
\usepackage[hyphens]{url}  
\usepackage{graphicx} 
\usepackage{natbib}  
\usepackage{caption} 
\usepackage{algorithm}
\usepackage{algorithmic}
\usepackage{booktabs}

\usepackage{newfloat}
\usepackage{listings}
\DeclareCaptionStyle{ruled}{labelfont=normalfont,labelsep=colon,strut=off} 
\floatstyle{ruled}
\newfloat{listing}{tb}{lst}{}
\floatname{listing}{Listing}
\usepackage{amssymb}
\usepackage{amsmath}
\usepackage{makecell}
\usepackage{color}
\usepackage{multirow}

\DeclareMathOperator*{\argmin}{arg\,min}

\title{Self-Navigated Residual Mamba for Universal Industrial Anomaly Detection}
\author{
    Hanxi Li\textsuperscript{\rm 1}\equalcontrib, Jingqi Wu\textsuperscript{\rm 2}\equalcontrib, Lin Yuanbo Wu\textsuperscript{\rm 3}, Mingliang Li\textsuperscript{\rm 1},\\
    Deyin Liu\textsuperscript{\rm 4}, Jialie Shen\textsuperscript{\rm 5}, Chunhua Shen\textsuperscript{\rm 6}
}
\affiliations{
   \textsuperscript{\rm 1}{Jiangxi Normal University,  China} \quad
     \textsuperscript{\rm 2}{Southern University of Science and Technology} \\
    \textsuperscript{\rm 3}{Swansea University, United Kingdom}\quad
 \textsuperscript{\rm 4}{Anhui University, China} \\
 \textsuperscript{\rm 5}{London University, United Kingdom}\quad
 \textsuperscript{\rm 6}{Zhejiang University, China}

}

\usepackage{bibentry}

\begin{document}

\maketitle

%
\begin{abstract}
  In this paper, we propose Self-Navigated Residual Mamba (SNARM), a novel framework for
  universal industrial anomaly detection that leverages ``self-referential
  learning'' within test images to enhance anomaly discrimination. Unlike conventional
  methods that depend solely on pre-trained features from normal training data, SNARM
  dynamically refines anomaly detection by iteratively comparing test patches against
  adaptively selected in-image references. Specifically, we first compute the
  ``inter-residuals'' features by contrasting test image patches with the training feature bank. Patches exhibiting small-norm residuals (indicating high
  normality) are then utilized as self-generated reference patches to compute ``intra-residuals'', amplifying discriminative signals. These inter- and intra-residual features are concatenated and fed into a novel Mamba module with multiple
  heads, which are dynamically navigated by residual properties to focus on anomalous
  regions. Finally, AD results are obtained by aggregating the outputs of a self-navigated Mamba in an ensemble learning paradigm. Extensive experiments on MVTec AD, MVTec 3D, and
  VisA benchmarks demonstrate that SNARM achieves state-of-the-art (SOTA) performance,
  with notable improvements in all metrics, including Image-AUROC,
  Pixel-AURC, PRO, and AP. The source code are available in
  \url{https://github.com/BeJane/SNARM.git}

\end{abstract}
\section{Introduction}
\label{sec:intro}

Industrial Anomaly Detection (IAD) is crucial in modern manufacturing, ensuring product
quality and reducing defects in automated production lines. However, the growing
complexity of industrial processes poses significant challenges. Production lines
increasingly handle diverse products, making the conventional category-specific anomaly
detection (AD) models \cite{roth2022towards, zavrtanik2021draem, li2023target} impractical
due to GPU memory and computational constraints. Therefore, the multi-class IAD problem
has attracted significant research attention \cite{liu2021swin, guo2024dinomaly,
he2024diffusion}. Additionally, the scarcity of training samples for every product
category has spurred interest in few-shot IAD algorithms \cite{gu_anomalygpt_2023,
jeong2023winclip, damm2025anomalydino}. To address these challenges more elegantly,
researchers have recently pursued a unified framework—termed Universal Anomaly Detection
(Univ-AD) \cite{luo2025exploring}—which shows promising scalability and efficiency for
real-world IAD tasks.

\begin{figure}[t!]  
\includegraphics [width=0.45\textwidth]{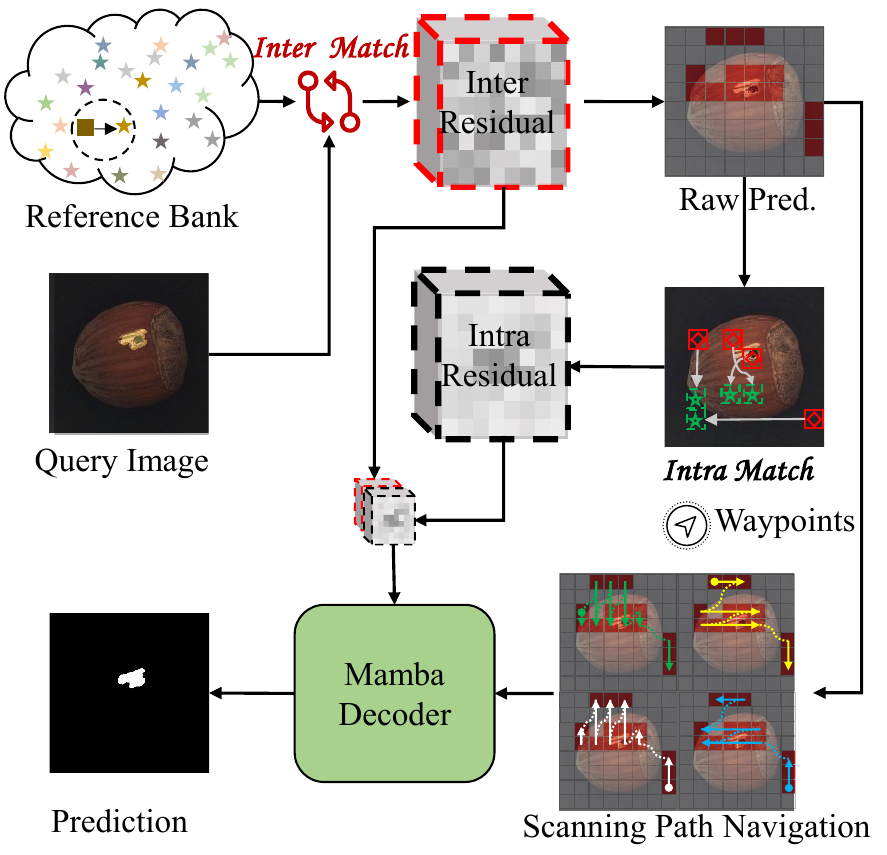}
\caption{
     Illustration of the proposed Self-Navigated Residual Mamba (SNARM).
    Different from conventional residual-based AD methods, the patch matching in this work contains two stages: the inter-matching stage and the intra-matching stage. The raw-prediction, \emph{i.e.}, the anomaly prediction based on the inter-matching guides the following intra-matching stage as well as the waypoint generation for navigating the scanning paths of the Mamba decoder.}
  \label{fig:concept}
\end{figure}

Recent advancements in deep learning have demonstrated the potential for Univ-AD tasks.
Some recently-proposed IAD methods \cite{li2023industrial, guo2024dinomaly,
damm2025anomalydino, luo2025exploring} leverage the modern architectures such Vision
Transformers (ViT) \cite{zhang2023exploring} to capture long-range dependencies and
hierarchical features for anomaly discrimination. On the contrary, as a well-acknowledged
competitor of Transformers, Mamba \cite{gu2023mamba, zhu2024vision} can generate
comparable performance while at a lower computational cost. However, the Mamba-based IAD
algorithms \cite{he2024mambaad, iqbal2025pyramid} can hardly beat the Transformer based
methods on most benchmarks.  Customizing the Mamba framework for industrial AD remains an
open challenge, particularly in refining feature representations and optimizing
computational efficiency.

In this paper, we propose \textbf{Self-Navigated Residual Mamba (SNARM)}, a novel Univ-AD algorithm that achieves state-of-the-art performance through three core innovations. First, instead of relying on a conventional one-stage patch-matching scheme, SNARM adopts a two-stage residual matching process. In the first stage, test patches are matched against a memory bank built from anomaly-free images to generate initial residuals. In the second stage, these test patches are re-matched against low-confidence patches—those predicted as normal by the first stage—producing more refined second-order residuals. This self-referential procedure is termed \textbf{Intra-Matching}, while the initial inter-image matching is referred to as \textbf{Inter-Matching}. Together, they form a \textbf{Hybrid Matching} strategy that enhances detection accuracy and reduces memory bank size, improving overall efficiency. Second, we introduce a \textbf{Self-Navigated Mamba Module} tailored for anomaly localization. Its scanning paths are dynamically guided by the residual properties, enabling more effective focus on anomalous regions. 
Finally, we aggregate the outputs from multiple directional and scale-specific Mamba branches using an ensemble learning framework, enhancing the robustness and reliability of predictions. An overview of the proposed method, compared with conventional residual-based approaches, is illustrated in Fig.~\ref{fig:concept}.

\vspace{0.5em}
\noindent\textbf{Our main contributions are summarized as follows:}
\begin{itemize}
  \item We propose \textbf{Hybrid Matching} for Univ-AD, where test patches participate in reference generation via self-referential matching, boosting both accuracy and efficiency.
  \item We design a \textbf{Self-Navigated Mamba Module} to enhance spatial focus on anomalies, and propose an ensemble-based decoder for robust multi-view integration.
  \item SNARM achieves state-of-the-art results on major benchmarks while maintaining high computational efficiency, making it suitable for real-world deployment.
\end{itemize}


\section{Related Work}
\label{sec:related}
\subsection{Industrial Anomaly Detection}
The field of Industrial Anomaly Detection (IAD) has seen rapid advancements since the
pioneering work~\cite{venkataramanan2020attention, roth2022towards, chen2022deep} laid the foundation for modern IAD algorithms. Early
research primarily focused on the single-class setting, where a dedicated model is trained
for each product category. The representative works include CutPaste~\cite{li2021cutpaste}, which leverages
self-supervised learning for anomaly localization, DRAEM~\cite{zavrtanik2021draem} with its discriminative
reconstruction embedding, and DRA~\cite{ding2022catching}, which learns a disentangled representation for unseen
anomalies. Other notable contributions, such as DeSTSeg~\cite{zhang2023destseg}, SimpleNet~\cite{Liu_2023_CVPR}, and RealNet~\cite{zhang2024realnet},
further advanced single-class AD through techniques like segmentation-guided denoising,
lightweight architectures, and realistic synthetic anomalies. As real-world industrial
scenarios often involve multiple product categories, researchers shifted toward
multi-class AD frameworks. Methods like works~\cite{liu2021swin, guo2024dinomaly, he2024diffusion} introduced scalable
solutions for handling diverse categories within a single model, while exploring plain ViT~\cite{zhang2023exploring} features demonstrated the potential of Vision Transformers in this setting. Another
critical challenge is the scarcity of labeled anomalies, motivating research in few-shot
AD. Works such as ~\cite{huang2022registration, gu_anomalygpt_2023, jeong2023winclip, damm2025anomalydino} explored ways to detect anomalies with minimal supervision, either through
visual or cross-modal cues. At the end of this spectrum lies universal AD,
exemplified by the algorithm INP-Former~\cite{luo2025exploring}, which aims to unify single-class, multi-class,
and few-shot AD under a single framework. In this paper, our work builds upon these
advancements to propose a more robust and scalable universal AD solution.

\subsection{Mamba-based IAD algorithms}
The recently proposed Mamba architecture~\cite{gu2023mamba}, inspired by State Space Models~\cite{gu2021efficiently}, has
emerged as a competitive alternative to Transformer-based models, offering comparable
performance with significantly lower computational complexity. This efficiency has spurred
the development of numerous Mamba-based visual models~\cite{liu2024vmamba, pei2024efficientvmamba, wang2024mamba, zhu2024vision, yang2025smamba} tailored for various downstream
tasks, including Industrial Anomaly Detection (IAD). For instance, Pyramid~\cite{iqbal2025pyramid}
adopts a Mamba model combined with a Pyramid Scanning Strategy to extract hierarchical
features from CNN representations. Meanwhile, MambaAD~\cite{he2024mambaad} introduces a Hybrid State Space
block to integrate global context for anomaly detection while leveraging local features
through traditional CNN modules. Although these methods demonstrate promising results in
multi-class IAD, they still fall short of surpassing the performance achieved by advanced
Transformer-based AD algorithms such as INP-Former~\cite{luo2025exploring} and Dinomaly~\cite{guo2024dinomaly}. In this work, we
present a meticulously designed Mamba framework specifically optimized for universal IAD,
achieving state-of-the-art performance across single-class, multi-class, and few-shot
anomaly detection scenarios.

\section{Method}
\label{sec:method}

\begin{figure*}[ht!]
  \centering
  \includegraphics[width=0.98\textwidth]{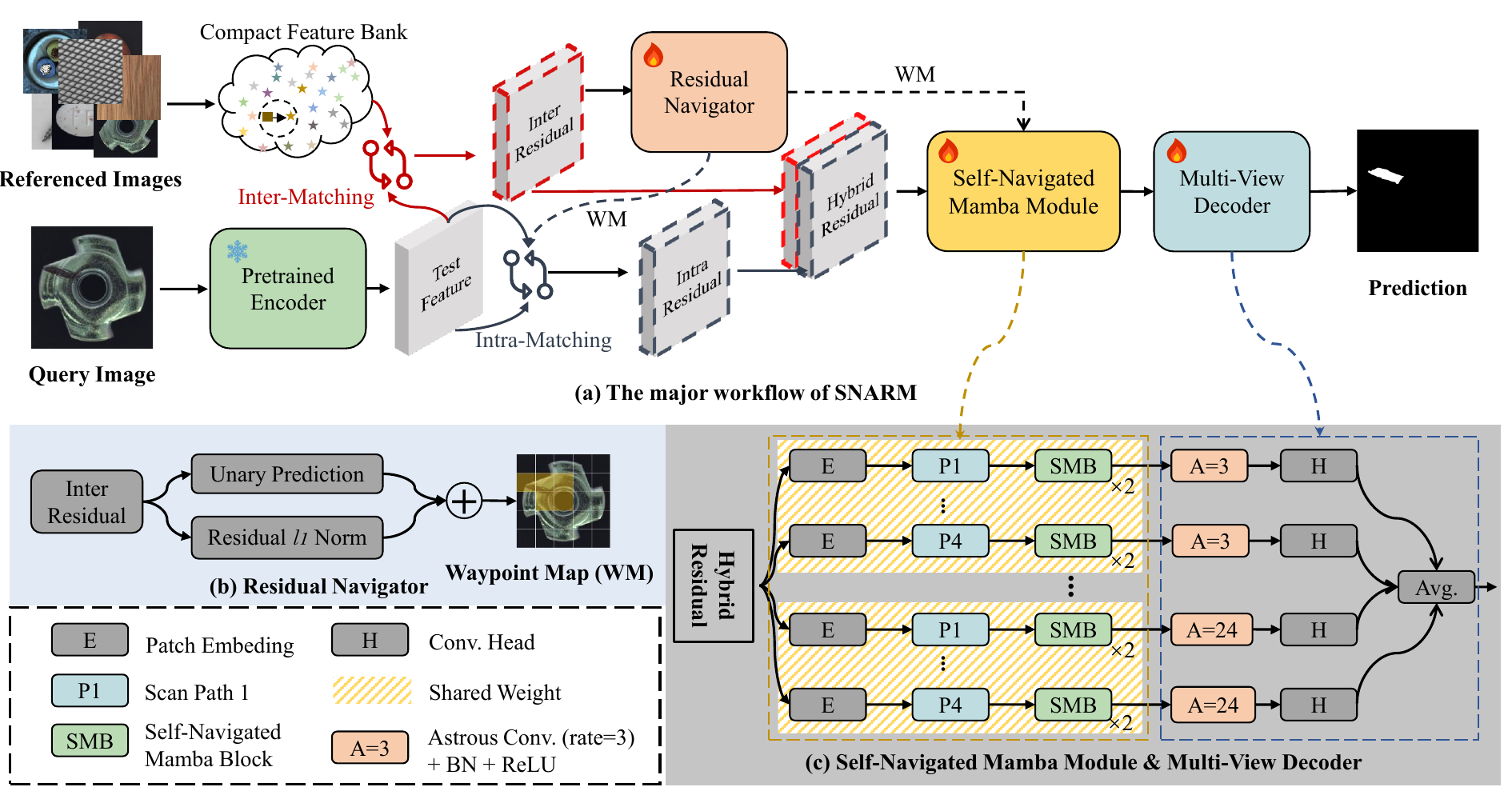}
   \caption{Overview of the proposed SNARM method. The query image is matched with an external feature bank to generate Inter-Residuals, followed by Inter-Matching to obtain Intra-Residuals. These are concatenated into Hybrid Residuals. A Waypoint Map (WM), derived from Inter-Residuals, dynamically guides the Self-Navigated Mamba Module to localize anomalies. Better viewed in color.}
  \label{fig:method}
\end{figure*}
\subsection{Overview}
\label{sec:overview}

We propose a novel residual-guided anomaly localization framework that enhances abnormality discrimination by integrating hybrid residual representations with dynamically guided Mamba branches, as illustrated in Fig.~\ref{fig:method}. The framework comprises four key components: \textbf{Hybrid Matching (Hybrid-M)}, a \textbf{Self-Navigated Mamba Module (SNMM)}, a \textbf{Multi-View Decoder (MVD)}, and an \textbf{ensemble learning strategy}. 

\subsection{Hybrid Matching}
\label{sec:matching}

We introduce a \textbf{Hybrid Matching (Hybrid-M)} strategy that integrates both \textit{Inter-Matching} and \textit{Intra-Matching}, guided by a Residual Navigator to effectively capture complementary normality patterns across images and within the test image itself.

\subsubsection{Universal Visual Feature Extraction}
\label{sec:feature_extraction}

Inspired by the potential of foundation models (e.g., DINOv2-R~\cite{darcet2023vision}) to provide general-purpose visual representations, we extract deep features from the input image $\mathrm{I} \in \mathbb{R}^{h_{\mathrm{I}} \times w_{\mathrm{I}} \times 3}$ using a pretrained encoder $\Psi_{\mathrm{EN}}(\cdot)$:
\begin{equation}
  \label{equ:extract}
  \begin{split}
    &\{\mathbf{F}^1, \mathbf{F}^2, \cdots, \mathbf{F}^L \} = \Psi_{\mathrm{EN}}(\mathrm{I}) \\
    &\mathbf{F} = \mathrm{Concat}\left(\frac{2}{L}\sum_{l=1}^{L/2}\mathbf{F}^l, \frac{2}{L}\sum_{l=L/2+1}^{L}\mathbf{F}^l\right) \\
    &[\mathbf{f}_1, \mathbf{f}_2, \cdots, \mathbf{f}_M] = \mathrm{Flatten}(\mathrm{Up}(\mathbf{F}))
  \end{split}
\end{equation}
Here, $\mathbf{F}^l \in \mathbb{R}^{h_f \times w_f \times d_f}$ denote the feature map from the $l$-th encoder block. We compute the average of early-layer (low-level) and late-layer (semantic) features, concatenate them, and then apply adaptive pooling following \cite{roth2022towards} to obtain a unified representation $\mathbf{F} \in \mathbb{R}^{h_f \times w_f \times d_f}$. This fused feature map is then upsampled and flattened into $M = h_f \cdot w_f$ local descriptors, denoted as ${\mathbf{f}_i \in \mathbb{R}^{d_f}}$.

\subsubsection{Inter-Matching}
\label{sec:ext_matching}

This Module aims to provide a consistent external reference for each test patch by querying a shared prototype bank. To construct this bank, we extract $M$ patch features from each normal training image and collect them into a raw feature pool:
\begin{equation}
\label{equ:raw_bank}
  \mathcal{B}_{\text{raw}} = \left\{\mathbf{f}^{\text{ref}}_{i,j} \in \mathbb{R}^{d_f} \mid j = 1, \cdots, N_{\text{trn}},~ i = 1, \cdots, M \right\},
\end{equation}
where $N_{\text{trn}}$ is the number of training images. To reduce memory and computation costs, we apply a coreset sampling algorithm~\cite{roth2022towards} to obtain a compact prototype bank:
\begin{equation} 
  \label{equ:coreset}
  \mathcal{B} = \Psi_{\mathrm{core}}(\mathcal{B}_{\text{raw}}) = \left\{\mathbf{f}^{\text{ref}}_t \in \mathbb{R}^{d_f} \mid t = 1, \cdots, T \right\},
\end{equation}
where $T \ll M \cdot N_{\text{trn}}$ and $\Psi_{\mathrm{core}}(\cdot)$ denotes the coreset selection function.

During inference, each test patch feature $\mathbf{f}^{\text{tst}}_i$ is matched with its nearest reference in the prototype bank using $L_2$ distance:
\begin{equation}
\label{equ:nearest}
t^{\ast} = \argmin_{t = 1, \cdots, T} \left\| \mathbf{f}^{\text{tst}}_i -
{\mathbf{f}}^{\text{ref}}_t \right\|_2.
\end{equation}
We then compute each patch of the Inter-Residual features $\mathbf{R} \in \mathbb{R}^{h_f \times w_f \times d_f} $ as a powered absolute difference:
\begin{equation}
\label{equ:residual}
\mathbf{r}_i = \left( \text{ABS}(\mathbf{f}^{\text{tst}}_i - \mathbf{f}^{\text{ref}}_{t^{\ast}}) \right)^{\theta}, \quad \forall i,
\end{equation}
where $\text{ABS}(\cdot)$ denotes the element-wise absolute difference, and the exponent $\theta \in \{1, 2\}$ is used to adjust the anomaly contrast level. 

\subsubsection{Intra-Matching Guided by Residual Navigator}
\label{sec:self-matching}

To improve matching fidelity and better highlight subtle anomalies, our method adaptively derives normal references from the test image itself. Unlike SoftPatch~\cite{jiang2022softpatch} and FunAD~\cite{im2025fun}, which leverage test features mainly for label denoising, or INP-Former~\cite{luo2025exploring}, which lacks explicit control over matching granularity, SNARM introduces a dedicated and interpretable matching framework that explicitly separates and fuses global and local cues, enabling more accurate and robust anomaly localization.

Specifically, we introduce a \textbf{Residual Navigator} that produces a coarse anomaly confidence map, termed the \textbf{Waypoint Map (WM)}, which highlights spatial regions likely to be abnormal. These pseudo-reference anchors are then used to recalibrate the residual distribution and guide downstream modules to focus on informative areas.

Given the inter-residual features $\mathbf{R} \in \mathbb{R}^{h_f \times w_f \times d_f}$, the Residual Navigator consists of two parallel branches:
\begin{equation}
\label{equ:navmap}
\mathbf{Q} = \sigma(\text{Conv}_{1\times1}(\mathbf{R})), \quad \mathbf{Q}^\star = \mathbf{Q} + \frac{1}{d_f} \sum_{k=1}^{d_f} \mathbf{R}(i,j,k),
\end{equation}
where $\text{Conv}_{1\times1}$ is a unary anomaly classifier composed of a $1 \times 1$ convolution followed by a sigmoid activation $\sigma(\cdot)$, and the second term computes the channel-wise average residual magnitude at each spatial location. The resulting $\mathbf{Q}^\star$ serves as the final Waypoint Map.
 
Then, we select a subset of reliable patches from the test image whose waypoint scores are among the lowest $p\%$. These low-score patches are assumed to be normal and are used as internal references for matching:
\begin{equation}
\label{equ:refined_bank}
\mathcal{S} = \left\{ \mathbf{f}^{\text{tst}}_k \,\middle|\, \mathbf{Q}^\star(k) < \text{Percentile}(\mathbf{Q}^\star, p) \right\},
\end{equation}
where $\mathbf{Q}^\star(k)$ denotes the waypoint score of the $k$-th patch, and $\text{Percentile}(\cdot, p)$ returns the $p$-th percentile threshold across all test patches.

Each patch $\mathbf{f}^{\text{tst}}_i$ is re-matched with the closest trusted patch $\mathbf{f}\in \mathcal{S}$ to compute the \textbf{Intra-Residuals} $\tilde{\mathbf{R}} \in \mathbb{R}^{h_f \times w_f \times d_f} $:
\begin{equation}
\label{equ:second-order}
\tilde{t}^{\ast} = \argmin_{\mathbf{f} \in \mathcal{S}} \left\| \mathbf{f}^{\text{tst}}_i - \mathbf{f} \right\|_2,
\quad
\tilde{\mathbf{r}}_i = \left( \text{ABS}(\mathbf{f}^{\text{tst}}_i - \mathbf{f}_{\tilde{t}^{\ast}}) \right)^{\theta}.
\end{equation}
Finally, we concatenate the inter-resiuals and intra-residuals to form the \textbf{Hybrid Residuals} $\hat{\mathbf{R}} \in \mathbb{R}^{h_f \times w_f \times 2d_f} $:
\begin{equation}
\label{equ:concat}
\hat{\mathbf{R}} = [\mathbf{R} \| {\tilde{\mathbf{R}}}],
\end{equation}
which encodes both global abnormality deviation and refined self-referenced differences.


\subsection{Self-Navigated Mamba Module}
\label{sec:snmm}

Mamba~\cite{gu2023mamba} has shown strong potential for capturing long-range dependencies via state-space modeling. However, directly applying it to dense anomaly localization is computationally expensive and less effective due to the large number of spatial tokens.

\begin{figure}[t]
  \centering
  \includegraphics[width=0.5\textwidth]{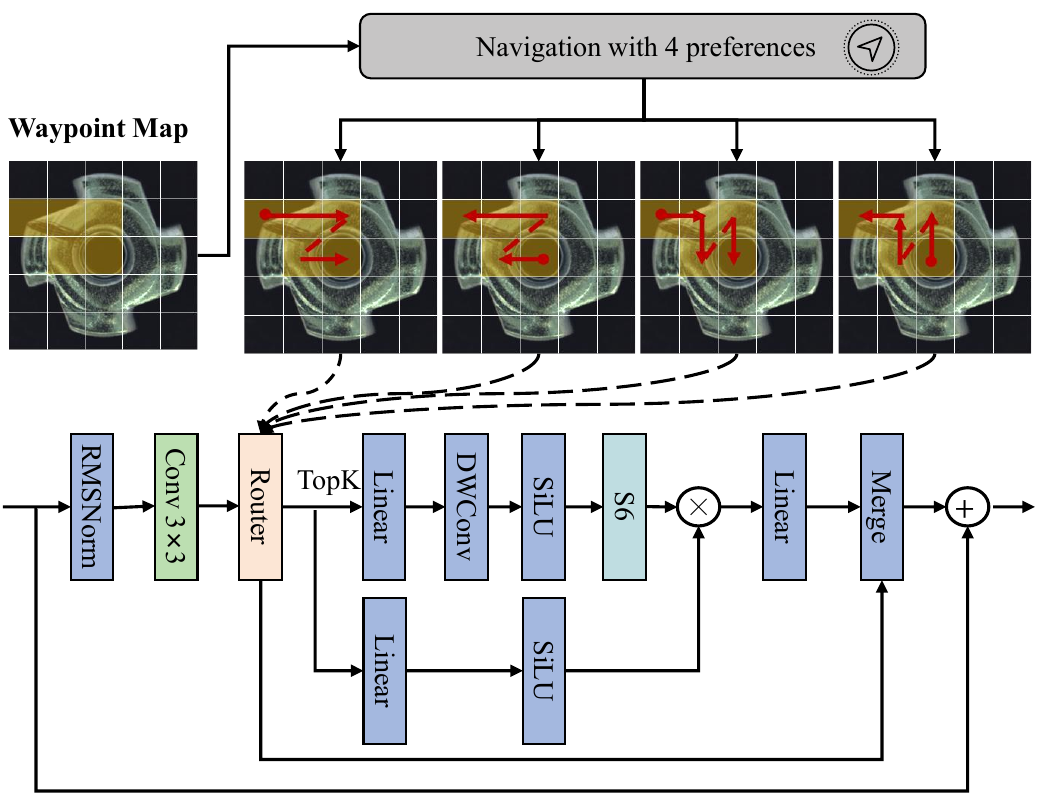}
   \caption{Overview of the proposed \textbf{Self-Navigated Mamba Block (SMB)}. Guided by the Waypoint Map, the block selectively scans informative regions using four directional passes, enabling efficient and spatially-aware message propagation over reduced token sets. Better viewed in color.}
  \label{fig:block}
\end{figure}

To address this, we propose the \textbf{Self-Navigated Mamba Block (SMB)}, which dynamically selects a compact and informative subset of tokens based on the Waypoint Map, as shown in Fig.~\ref{fig:block}. To enhance spatial perception, SMB performs directional scanning in four orientations—left-to-right, right-to-left, top-down, and bottom-up—allowing better contextual aggregation across axes. To further enrich local context and propagate anomaly cues, a $3 \times 3$ convolutional layer is placed before each scan.

The complete \textbf{Self-Navigated Mamba Module (SNMM)} consists of a feature embedding layer followed by two stacked SMBs. Given the hybrid-residual features, SNMM generates four directionally-aware outputs $\{\mathbf{o}^{(\rightarrow)}, \mathbf{o}^{(\leftarrow)}, \mathbf{o}^{(\downarrow)}, \mathbf{o}^{(\uparrow)}\}$, which are passed to the decoder for final anomaly prediction.

\subsection{Multi-View Decoder}
\label{sec:mvd}

In IAD, anomalies vary significantly in scale, shape, and contextual appearance—ranging from subtle scratches to large structural defects. To robustly capture such heterogeneous patterns, we introduce a \textbf{Multi-View Decoder (MVD)}, composed of multiple \textbf{View-Specific Branches}, each tailored to a distinct spatial perspective (multi-scan and multi-scale), as illustrated in Fig.~\ref{fig:method}(c).

Each branch receives a directional feature map $\mathbf{o}^{(d)}$ from the SNMM, where $d \in \{\rightarrow, \leftarrow, \downarrow, \uparrow\}$ denotes the scan direction. To enhance adaptability to varying anomaly scales, we employ a branch-specific Atrous Convolution Block $\mathcal{A}_{r,d}(\cdot)$ with a unique dilation rate $r \in \{3, 6, 12, 24\}$:
\begin{equation}
\label{eq:atrous}
\tilde{\mathbf{o}}^{(r,d)} = \mathcal{A}_{r,d}\left(\mathbf{o}^{(d)}\right) = \text{AtrousConv}\left(\mathbf{o}^{(d)}; r\right).
\end{equation}
The transformed features are then passed through a lightweight prediction head $\mathcal{H}(\cdot)$, which comprises a $1 \times 1$ convolution, an upsampling operation, and a Sigmoid activation $\sigma(\cdot)$ to produce the anomaly map:
\begin{equation}
\label{eq:head}
\mathbf{m}^{(r,d)} = \mathcal{H}(\tilde{\mathbf{o}}^{(r,d)}) = \sigma\left(\text{Up}\left(\text{Conv}_{1 \times 1}(\tilde{\mathbf{o}}^{(r,d)})\right)\right),
\end{equation}
where $\mathbf{m}^{(r,d)}$ denotes the anomaly probability map corresponding to view $d$ and dilation rate $r$.

\subsection{Ensemble Learning Strategy}
\label{sec:ensemble}

\subsubsection{Cyclic Optimization and Loss Function}
\label{sec:cyclic}
An iterative training strategy is adopted to promote diversity among the  \textit{View-Specific Branches}. Specifically, we sequentially update one branch at a time for $\mathcal{K}$ steps while freezing the others. 

We adopt the Focal Loss \cite{lin2017focal} to supervise both the Residual Navigator and the Multi-View Decoder. During each cycle, only the active branch contributes to the training loss, while the Residual Navigator is jointly optimized across the entire training process. The total loss $\mathcal{L}_{\text{total}}$ at each training iteration is as follow:
\begin{equation}
\label{equ:total_loss}
\mathcal{L}_{\text{total}} = \mathcal{L}_{\text{focal}}(\text{Up}(\mathbf{Q}), \mathbf{Y}) + \frac{1}{4} \sum_{j=1}^{4} \mathcal{L}_{\text{focal}}(\mathbf{m}^{i,j}, \mathbf{Y}),
\end{equation}
where $\text{Up}(\cdot)$ denotes a bilinear interpolation operation that resizes the low-resolution output $\mathbf{Q}$ from the Residual Navigator's convolution branch in Equ.~\ref{equ:navmap} to the original input resolution of the ground truth map  $\mathbf{Y} \in \{0,1\}^{h_I \times w_I}$. $\mathbf{m}^{i,j}$ is the anomaly prediction from the $j$-th directional sub-branch of the $i$-th scale-specific branch, and $\mathcal{L}_{\text{focal}}(\cdot)$ is the Focal Loss~\cite{lin2017focal} defined as:
\begin{equation}
\begin{aligned}
\label{equ:focal_loss}
 \mathcal{L}_{\text{focal}}(\hat{\mathbf{M}}, \mathbf{Y}) = - \sum_{x,y} [ \alpha (1 - \hat{\mathbf{M}}_{x,y})^{\gamma} \mathbf{Y}_{x,y} \log(\hat{\mathbf{M}}_{x,y})  \\
 + (1 - \alpha) \hat{\mathbf{M}}_{x,y}^{\gamma} (1 - \mathbf{Y}_{x,y}) \log(1 - \hat{\mathbf{M}}_{x,y}) ],   
 \end{aligned}
\end{equation}
where $\hat{\mathbf{M}}_{x,y}$ denotes the predicted probability at pixel location $(x, y)$, $\alpha$ is the weighting factor for class imbalance, and $\gamma$ is the focusing parameter to down-weight easy examples. 

\subsubsection{Feature Augmentation Strategies}
\label{sec:feature_aug}

To improve the robustness and generalization of the model, we introduce two feature augmentation strategies tailored for residual-based anomaly detection.

\textbf{(1) Top-$k$ Feature Averaging}  
In conventional residual computation, $\mathbf{f}_{t^{\ast}}^{\text{ref}}$ is the most similar reference feature from the memory bank $\mathcal{B}$ when computing the Inter-Residual features according to Equ.~\ref{equ:residual}. We extend this by averaging the top-$k$ nearest features $\{ \mathbf{f}_{t_{1}}, \dots, \mathbf{f}_{t_k} \}$ to form a more stable reference:
\begin{equation}
\label{equ:topk}
\mathbf{f}_i^{\text{ref}} = \frac{1}{k} \sum_{j=1}^{k} \mathbf{f}_{t_j}.
\end{equation}
This approach smooths residual computation, mitigating the effect of noise. It can be applied during both training and inference stages.

\textbf{(2) Consistent Feature Jittering}  
Inspired by Feature Jittering in UniAD~\cite{you2022unified}, we introduce a \textit{noise-consistent} augmentation mechanism. Specifically, we sample a noise vector $\boldsymbol{\epsilon} \sim \mathcal{N}(0, \mathbf{I})$ for each sample and add it to both its universal visual feature $\mathbf{f}_i$ and its Inter-Residual feature $\mathbf{r}_i$, ensuring consistency in perturbation:
\begin{equation}
\label{equ:jitter}
\tilde{\mathbf{f}}_i = \mathbf{f}_i + \lambda \frac{\left\|\mathbf{f}_i\right\|_2}{d_f}\boldsymbol{\epsilon}, \quad
\tilde{\mathbf{r}}_i = \mathbf{r}_i + \lambda \frac{\left\|\mathbf{r}_i\right\|_2}{d_f}\boldsymbol{\epsilon} \quad \forall i.
\end{equation}
where $\lambda$ is the jittering scale to control the noisy degree.

\subsubsection{Inference}
\label{sec:infer}
The final anomaly map is obtained by averaging the outputs from all $4 \times 4$ view-specific branches:
\begin{equation}
\hat{\mathbf{m}} = \frac{1}{16}\sum_{i=1}^{4} \sum_{j=1}^{4} \mathbf{m}^{i,j},
\end{equation}
where $\hat{\mathbf{m}} \in [0,1]^{h_I \times w_I}$ denotes the aggregated confidence map, and $\mathbf{m}^{i,j}$ denotes the anomaly prediction from the $j$-th directional sub-branch of the $i$-th scale-specific branch. 

\subsection{Implementation Details}
\label{detail}

SNARM employs ViT-Base/14 with DINOv2-R~\cite{darcet2023vision} weights as the default pre-trained encoder $\Psi_{\mathrm{EN}}(\cdot)$. Feature maps are extracted from the intermediate outputs of the 1st to 8th transformer blocks. Input images are first resized to $448 \times 448$ and then center-cropped to $392 \times 392$. The training hyperparameters are set as follows: $T = 10^4$ in Eq.\ref{equ:coreset}, $p = 75$ in Eq.\ref{equ:refined_bank}, $k = 3$ in Eq.~\ref{equ:topk}, $\mathcal{K} = 100$, and $\lambda = 30$. For the Residual Navigator loss, we use $\alpha = 0.5$ and $\gamma = 4$, while for the View-Specific Branch loss, we set $\alpha = 0.25$ and $\gamma = 4$. The entire framework is optimized using the Adam optimizer with a learning rate of $0.001$ and a weight decay of $0.05$. All experiments are conducted on a single PC equipped with an Intel i5-13450 CPU, 64GB RAM, and an NVIDIA RTX 4090 GPU.

\section{Experiments}
\label{sec:experiments}
\begin{table*}[!htb]

\centering
\resizebox{\linewidth}{!}{
\begin{tabular}{l|cccc|cccc|cccc}
\toprule
       Dataset               & \multicolumn{4}{c|}{MVTec-AD~\cite{bergmann2019mvtec}}                   & \multicolumn{4}{c|}{MVTec-3D~\cite{bergmann2021mvtec}}                   & \multicolumn{4}{c}{VisA~\cite{zou_spot--difference_2022}}                       \\
       \midrule
Method                & P\_AP & PRO  & P\_AUROC & I\_AUROC & P\_AP & PRO  & P\_AUROC & I\_AUROC & P\_AP & PRO  & P\_AUROC & I\_AUROC \\
\midrule
RD4AD~\cite{deng2022anomaly}& 48.6      & 91.1 & 96.1         & 94.6         & 29.8      & 93.5 & 98.4         & 77.9         & 38.0      & 91.8 & 98.1         & 92.4         \\
UniAD~\cite{you2022unified}  & 43.4      & 90.7 & 96.8         & 96.5         & 21.2      & 88.1 & 96.5         & 78.9         & 33.7      & 85.5 & 98.3         & 88.8         \\
SimpleNet~\cite{Liu_2023_CVPR}& 45.9      & 86.5 & 96.9         & 95.3         & 18.3      & 77.6 & 93.5         & 72.5         & 34.7      & 81.4 & 96.8         & 87.2         \\
DeSTSeg~\cite{zhang2023destseg}& 54.3      & 64.8 & 93.1         & 89.2         & 38.1      & 46.4 & 95.1         & 79.6         & 39.6      & 67.4 & 96.1         & 88.9         \\
DiAD~\cite{he2024diffusion} & 52.6      & 90.7 & 96.8         & 97.2         & 25.3      & 87.8 & 96.4         & 84.6         & 26.1      & 75.2 & 96.0         & 86.8         \\
MambaAD~\cite{he2024mambaad}  & 56.3      & 93.1 & 97.7         & 98.6         & 37.5      & 93.6 & 98.6         & 86.2         & 39.4      & 91.0 & 98.5         & 94.3         \\
WeakREST~\cite{li2023industrial}              & \textbf{\color{blue}77.1}      & 95.4 & 98.3         & 98.5         & 48.8      & 95.4 & 98.6         & 83.8         &  $\sim$         &   $\sim$     &        $\sim$        &    $\sim$                   \\
CPR \cite{li2023target}                  & 63.3      & 93.1 & 97.2         & 95.7         & 37.6      & 95.2 & 98.4         & 80.9         &  $\sim$         &   $\sim$     &        $\sim$        &    $\sim$            \\
Dinomaly~\cite{guo2024dinomaly}              & 68.7      & 94.7 & 98.3         & \textbf{\color{blue}99.6}         & 55.0      & 96.5 & \textbf{\color{red}99.2}         & 90.6         & 53.2      & \textbf{\color{blue}94.5} & 98.7         & \textbf{\color{blue}98.7}         \\
INP-Former~\cite{luo2025exploring}  & 71.0      & 94.9 & 98.5         & \textbf{\color{red}99.7}         & 55.9      & \textbf{\color{blue}97.0} & \textbf{\color{red}99.2}     & \textbf{\color{blue}92.6}         & 51.2      & 94.4 & \textbf{\color{blue}98.9}         & \textbf{\color{red}98.9}         \\ \midrule
\textbf{SNARM} ($T=10^4$)        & 76.9      & \textbf{\color{blue}95.9} & \textbf{\color{blue}98.9}         & 99.1         & \textbf{\color{blue}59.5}      & 96.5 & \textbf{\color{blue}99.0}         & 90.2         & \textbf{\color{blue}55.6}      & 90.1 & 98.4         & 96.8         \\
\textbf{SNARM} ($T=10^5$)                & \textbf{\color{red}79.0}      & \textbf{\color{red}96.6} & \textbf{\color{red}99.1}         & 99.4         & \textbf{\color{red}63.6}      & \textbf{\color{red}97.4} & \textbf{\color{red}99.2}         & \textbf{\color{red}93.9}         & \textbf{\color{red}55.8}      & \textbf{\color{red}94.7} & \textbf{\color{red}99.1}         & 98.1        \\\bottomrule
\end{tabular}}
\caption{
  \textbf{Multi-class} anomaly detection performance on different AD datasets. The best accuracy in one comparison with the same data and metric condition is shown in \textbf{\color{red}red} while the second one is shown in \textbf{\color{blue}blue}.
  }
\label{tab:multi_class}
\end{table*}

\begin{table*}[!tb]

\centering
\resizebox{\linewidth}{!}{
\begin{tabular}{l|cccc|cccc|cccc}
\toprule
       Dataset               & \multicolumn{4}{c|}{MVTec-AD~\cite{bergmann2019mvtec}}                   & \multicolumn{4}{c|}{MVTec-3D~\cite{bergmann2021mvtec}}                   & \multicolumn{4}{c}{VisA~\cite{zou_spot--difference_2022}}                       \\
       \midrule
Method                & P\_AP & PRO  & P\_AUROC & I\_AUROC & P\_AP & PRO  & P\_AUROC & I\_AUROC & P\_AP & PRO  & P\_AUROC & I\_AUROC \\
\midrule
PaDiM~\cite{defard2021padim}                            & 53.6  & 92.2 & 97.3     & 95.5         & 23.7  & 91.0 & 97.3     & 76.6         & 31.3  & 83.6 & 97.6     & 89.9         \\
PatchCore~\cite{roth2022towards}      & 57.1  & 93.1 & 98.1     & 99.1         & 26.3  & 89.1 & 97.0     & 82.5         & \textbf{\color{blue}39.5}  & 86.2 & \textbf{\color{blue}97.8}     & \textbf{\color{blue}90.4}         \\
CPR~\cite{li2023target}                              & 61.4  & 91.4 & 96.8     & 92.8         & 23.6  & 87.6 & 96.3     & 71.3         & 30.8  & 77.5 & 92.7     & 87.8         \\
HETMM~\cite{chen2024hard}                            & \textbf{\color{blue}68.0}  & \textbf{\color{blue}95.8} & \textbf{\color{blue}98.6}     & \textbf{\color{blue}99.2}         & \textbf{\color{blue}37.7}  & \textbf{\color{blue}94.8} & \textbf{\color{blue}98.4}     & \textbf{\color{blue}83.4}         & 38.4  & \textbf{\color{blue}89.1} & 96.3     & 89.0         \\
Dinomaly~\cite{guo2024dinomaly}                         & 15.2  & 39.5 & 73.8     & 63.6         & 4.2   & 59.5 & 87.8     & 46.9         & 9.8   & 51.5 & 86.3     & 59.5         \\ \midrule
\textbf{SNARM} ($T=10^4$) & \textbf{\color{red}78.4}  & \textbf{\color{red}96.1} & \textbf{\color{red}99.0}     & \textbf{\color{red}99.3}         & \textbf{\color{red}68.9}  & \textbf{\color{red}97.3} & \textbf{\color{red}99.1}     & \textbf{\color{red}94.5}         & \textbf{\color{red}56.0}  & \textbf{\color{red}93.8} & \textbf{\color{red}99.1}     & \textbf{\color{red}98.0} \\\bottomrule        
\end{tabular}}
\caption{
\textbf{Cross-class} anomaly detection performance on different AD datasets. 
  }
\label{tab:cross_class}
\end{table*}

\begin{table*}[hbt]

\centering
\resizebox{\linewidth}{!}{
\begin{tabular}{l|cccc|cccc|cccc}
\toprule
       Dataset               & \multicolumn{4}{c|}{MVTec-AD~\cite{bergmann2019mvtec}}                   & \multicolumn{4}{c|}{MVTec-3D~\cite{bergmann2021mvtec}}                   & \multicolumn{4}{c}{VisA~\cite{zou_spot--difference_2022}}                       \\
       \midrule
Method                & P\_AP & PRO  & P\_AUROC & I\_AUROC & P\_AP & PRO  & P\_AUROC & I\_AUROC & P\_AP & PRO  & P\_AUROC & I\_AUROC \\
\midrule
RD4AD~\cite{deng2022anomaly}      & 58.0  & 93.9 & 97.8     & 98.0   $\sim$      &     $\sim$   &   $\sim$   &  $\sim$        &  $\sim$            & 27.7  & 70.9 &          & 96.0         \\
SimpleNet~\cite{Liu_2023_CVPR}  & 54.8  & 90.0 & 97.7     & \textbf{\color{blue}99.6}         &    $\sim$   &     $\sim$ &     $\sim$     &  $\sim$            & 36.3  & 88.7 &          & 96.8         \\
Dinomaly~\cite{guo2024dinomaly}    & 68.9  & 95.0 &  $\sim$        & \textbf{\color{red}99.7}         &   $\sim$    &    $\sim$  &     $\sim$     &    $\sim$          & \textbf{\color{blue}50.7}  & \textbf{\color{red}95.1} &          & \textbf{\color{red}98.9}         \\
INP-Former~\cite{luo2025exploring} & \textbf{\color{blue}70.2}  & \textbf{\color{blue}95.4} & \textbf{\color{blue}98.3}     & \textbf{\color{red}99.7}         & \textbf{\color{blue}56.6}  & \textbf{\color{blue}97.0} & \textbf{\color{red}99.2}     & \textbf{\color{blue}93.0}         & 49.2  & \textbf{\color{blue}93.8} & \textbf{\color{blue}98.4}     & \textbf{\color{blue}98.5}         \\ \midrule
\textbf{SNARM} ($T=10^4$)  &     \textbf{\color{red}79.2}      & \textbf{\color{red}96.6} & \textbf{\color{red}99.0}         & 99.3         & \textbf{\color{red}65.5}  & \textbf{\color{red}97.2} & \textbf{\color{red}99.2}     & \textbf{\color{red}94.0}         & \textbf{\color{red}55.4}  & 92.3 & \textbf{\color{red}99.0}     & 98.0   \\\bottomrule     
\end{tabular}}
\caption{
\textbf{Single-class} anomaly detection performance on different AD datasets. 
  }
\label{tab:single_class}
\end{table*}

\begin{table*}[hbt]

\centering
\resizebox{0.8\linewidth}{!}{
\begin{tabular}{l|cccc|cccc}
\toprule
       Dataset               & \multicolumn{4}{c|}{MVTec-AD~\cite{bergmann2019mvtec}}                   & \multicolumn{4}{c}{VisA~\cite{zou_spot--difference_2022}}                       \\
       \midrule
Method                & P\_AP & PRO  & P\_AUROC & I\_AUROC & P\_AP & PRO  & P\_AUROC & I\_AUROC \\
\midrule
PaDiM\cite{defard2021padim}         &  $\sim$       & 81.3 & 92.6     & 80.4         &    $\sim$     & 72.6 & 93.2     & 72.8         \\
PatchCore~\cite{roth_towards_2022}   &    $\sim$     & 84.3 & 94.3     & 88.8         &     $\sim$    & 84.9 & 96.8     & 85.3         \\
WinCLIP~\cite{jeong2023winclip}     &   $\sim$      & 89.0 & 96.2     & 95.2         &  $\sim$       & 87.6 & 97.2     & 87.3         \\
PromptAD~\cite{promptad}    &   $\sim$      & 90.5 & 96.5     & 96.6         &   $\sim$      & 86.2 & 97.4     & 89.1         \\
AnomalyDINO~\cite{damm2025anomalydino} &      $\sim$   & 92.4 & 96.7     & \textbf{\color{blue}97.6}         &      $\sim$   & 92.5 & \textbf{\color{blue}98.0}     & 91.3         \\
INP-Former~\cite{luo2025exploring}  & \textbf{\color{blue}65.9}  & \textbf{\color{blue}92.9} & \textbf{\color{blue}97.0}     & \textbf{\color{blue}97.6}         & \textbf{\color{blue}49.3}  & \textbf{\color{red}93.1} & 97.7     & \textbf{\color{red}96.4}         \\\midrule
\textbf{SNARM} ($T=10^4$)    & \textbf{\color{red}68.6}  & \textbf{\color{red}94.6} & \textbf{\color{red}97.9}     & \textbf{\color{red}98.3}         & \textbf{\color{red}49.5}  & \textbf{\color{blue}92.9} & \textbf{\color{red}98.1}     & \textbf{\color{blue}94.5}      \\ \bottomrule  
\end{tabular}
}
\caption{
\textbf{Few-shot (4-shot)} anomaly detection performance on different AD datasets. 
  }
\label{tab:few_shot}
\end{table*}

\subsection{Experimental Setting}
The experiments are conducted on three widely-used benchmarks: the MVTec-AD~\cite{bergmann2019mvtec} MVTec- 3D~\cite{bergmann2021mvtec}  and VisA~\cite{zou_spot--difference_2022} datasets. To ensure a fair and comprehensive evaluation, we adopt four threshold-independent metrics: Image-level AUROC  (I-AUROC), Pixel-level AUROC (P-AUROC), Per-Region Overlap (PRO), and Average Precision (P-AP). Among them, I-AUROC assesses the accuracy of anomaly detection at the image level, while P-AUROC, PRO, and P-AP evaluate the performance of anomaly localization.

\subsection{Results under the Multi-Class Setting} 
\label{sec:multi_class}
As shown in Table~\ref{tab:multi_class}, SNARM achieves state-of-the-art or highly competitive results across all three datasets, even with a highly compact memory bank. On MVTec-AD~\cite{bergmann2019mvtec}, it achieves superior pixel-level performance (\textbf{79.0/96.6/99.1}) and a competitive I-AUROC, surpassing prior best results by \textbf{1.9/1.7/0.6}. On the more challenging MVTec-3D~\cite{bergmann2021mvtec}, SNARM achieves \textbf{63.6/97.4/99.2/93.9}, outperforming previous methods by margins of \textbf{7.7/0.4/0.0/1.3}. Similarly, on VisA~\cite{zou_spot--difference_2022}, it attains \textbf{55.8/94.7/99.1} with a competitive I-AUROC, improving over prior results by \textbf{2.6/0.2/10.2}. Notably, reducing the memory bank size has minimal impact on performance, highlighting the robustness and efficiency of our framework. Detailed per-class results are provided in the Supplementary Material. 

\subsection{Results under the Cross-Class Setting}
\label{sec:cross_class}

In this setting, models are evaluated on previously unseen classes without additional retraining. For embedding-based methods (e.g., PatchCore~\cite{roth2022towards}, PaDiM~\cite{defard2021padim},  HETMM~\cite{chen2024hard}), we employ a backbone pre-trained on a large-scale dataset such as ImageNet~\cite{russakovsky2015imagenet}. For learning-based methods (e.g., SNARM, Dinomaly~\cite{guo2024dinomaly} and CPR~\cite{li2023target}), we adopt a leave-one-out strategy, where one class is designated for testing while the remaining classes are used for training.

As shown in Table~\ref{tab:cross_class}, our method achieves best-in-class performance across all benchmarks. On MVTec-AD~\cite{bergmann2019mvtec}, it attains scores of \textbf{78.4/96.1/99.0/99.3}, outperforming the previous best by \textbf{10.4/0.3/0.4/0.1}, respectively. For MVTec-3D~\cite{bergmann2021mvtec}, our method reaches \textbf{68.9/97.3/99.1/94.5}, with margins of \textbf{31.2/2.5/0.7/11.1} over the prior state-of-the-art. Similarly, on VisA~\cite{zou_spot--difference_2022}, it achieves \textbf{56.0/93.8/99.1/98.0}, surpassing existing methods by \textbf{16.5/4.7/1.3/7.6}, further validating its strong generalization and effectiveness.  Detailed per-class results are provided in the Supplementary Material. 

\subsection{Results under the Single-Class Setting}
\label{sec:single_class}

As presented in Table~\ref{tab:single_class}, SNARM delivers state-of-the-art or highly competitive performance across all three benchmarks under the Single-class setting.  On MVTec-AD~\cite{bergmann2019mvtec}, SNARM achieves outstanding pixel-level scores of \textbf{79.2/96.6/99.0}, exceeding previous best results by \textbf{9.0/1.2/0.7}. On MVTec-3D~\cite{bergmann2021mvtec}, our method reaches \textbf{65.5/97.2/99.2/94.0}, surpassing prior approaches by margins of \textbf{8.9/0.2/0.0/1.0}.  On the VisA dataset~\cite{zou_spot--difference_2022}, SNARM also demonstrates competitive performance. Detailed per-category metrics are provided in the Supplementary Material.

\subsection{Results under the Few-Shot Setting} 
\label{sec:few_shot}
As shown in Table~\ref{tab:few_shot}, SNARM achieves state-of-the-art or highly competitive performance under the few-shot setting. On MVTec-AD~\cite{bergmann2019mvtec}, it obtains impressive scores of \textbf{68.6/94.6/97.9/98.3}, outperforming previous methods by margins of \textbf{2.7/1.7/0.9/0.7}. On the VisA dataset~\cite{zou_spot--difference_2022}, SNARM also achieves strong and competitive results.

\begin{table*}[!hbt]

\centering
\resizebox{0.8\linewidth}{!}{
\begin{tabular}{cccccc|cccc|cccc}\toprule
\multicolumn{6}{c|}{Module}                 & \multicolumn{4}{c|}{Multi-Class}                                                                                   & \multicolumn{4}{c}{Few-Shot}         \\ \midrule
Hybrid-M & SMB & MVD &  TopK & Jitter & CT& P\_AP & PRO  & P\_AUROC & I\_AUROC & P\_AP & PRO  & P\_AUROC & I\_AUROC \\ \midrule
    &      &     &    &      &        &     70.2                      & 93.8                    & 97.7                         & 99.0                         & 64.6                      & 93.9                    & 97.1                         & \multicolumn{1}{c}{97.71}       \\
\textbf{\checkmark}   &      &     &    &      &        &    71.5                      & 93.5                    & 97.9                         & 98.8                         & 65.9                      & 94.0                    & 97.3                         & 97.9                            \\
\textbf{\checkmark}& \textbf{\checkmark} &     &    &      &        &     73.8                      & 94.4                    & 98.3                         & 98.9                         & 66.3                      & 93.6                    & 96.6                         & 97.2                          \\
\textbf{\checkmark}& \textbf{\checkmark} & \textbf{\checkmark}& &     &        &     76.9                      & 95.3                    & 98.8                         & \textbf{\color{red}99.1}                         & 67.7                      & 93.5                    & 97.1                         & 96.8                             \\
\textbf{\checkmark}&  \textbf{\checkmark}& \textbf{\checkmark}& \textbf{\checkmark} &    &    &    \textbf{\color{red}77.5}                      & 95.7                    & 98.8                         & \textbf{\color{red}99.1}                         & 67.7                      & 93.4                    & 97.1                         & 97.0                            \\
\textbf{\checkmark}& \textbf{\checkmark} & \textbf{\checkmark}& \textbf{\checkmark}& \textbf{\checkmark} &    &    \textbf{\color{red}77.5}                      & 95.6                    & \textbf{\color{red}98.9 }                        & 99.0                         & 68.5                      & \textbf{\color{red}94.7}                    & \textbf{\color{red}97.9}                         & 98.1                           \\
\textbf{\checkmark}& \textbf{\checkmark} & \textbf{\checkmark}& \textbf{\checkmark}& \textbf{\checkmark} &  \textbf{\checkmark} &76.9                      & \textbf{\color{red}95.9}                    & \textbf{\color{red}98.9}                         & \textbf{\color{red}99.1}                  & \textbf{\color{red}68.6}                      & 94.6                    & \textbf{\color{red}97.9}                         & \textbf{\color{red}98.3}                        \\\bottomrule  
\end{tabular}}
\caption{
Ablation on the MVTec-AD~\cite{bergmann2019mvtec} dataset under the Multi-class and Few-shot  (4-shot) settings.
  }
\label{tab:ablation}
\end{table*}

\begin{table}[!htb]

\centering
\resizebox{\linewidth}{!}{
\begin{tabular}{ccccc}\toprule
Method     & Params (M) & FLOPs(G) & FPS & mAD  \\\midrule
DeSTSeg~\cite{zhang2023destseg}    & \textbf{\color{blue}35.2}       & 122.7    & \textbf{\color{red}125} & 75.4 \\
SimpleNet~\cite{Liu_2023_CVPR}  & 72.8       & \textbf{\color{blue}16.1}     & \textbf{\color{blue}62}  & 81.2 \\
MambaAD~\cite{he2024mambaad}    & \textbf{\color{red}25.7}       & \textbf{\color{red}8.3}      & 59  & 86.4 \\
Dinomaly~\cite{guo2024dinomaly}   & 132.8      & 104.7    & 59  & 90.3 \\
INP-Former~\cite{luo2025exploring} & 139.8      & 98.0       & 38  & \textbf{\color{blue}91.0} \\ \midrule
\textbf{SNARM} ($T=10^4$)   & 68.3       & 69.0       & 48  & \textbf{\color{red}92.7}\\\bottomrule
\end{tabular}}
\caption{
Comparison of computational efficiency among SOTA methods. mAD represents the average value of four metrics under Multi-class setting on the MVTec-AD~\cite{bergmann2019mvtec} dataset.
  }
\label{tab:speed}
\vspace{-0.3cm}
\end{table}

\subsection{Complexity Comparisons}
\label{sec:complexity}

Table~\ref{tab:speed} compares SNARM with recent state-of-the-art methods in terms of efficiency. SNARM achieves the best mAD score (92.7) with only 69 GFLOPs, 68.3M parameters, and competitive speed (48 FPS), offering an excellent trade-off between accuracy and complexity. 

\subsection{Ablation Study}
\label{sec:ablation}

We conduct an ablation study on the MVTec-AD dataset~\cite{bergmann2019mvtec} under multi-class and few-shot settings (Table~\ref{tab:ablation}). Starting from the base model with Inter-Residuals and vanilla Mamba, we first add \textbf{Hybrid-Matching (Hybrid-M)}, which significantly boosts performance by injecting informative Intra-Residuals. Integrating the \textbf{Self-Navigated Mamba Block (SMB)} further enhances anomaly localization via guided attention. Introducing the \textbf{Multi-View Decoder (MVD)}  strengthens spatial adaptability across varied view. Then, the \textbf{Top-$k$ Feature Averaging (TopK)} module improves robustness, the \textbf{Consistent Feature Jittering (Jitter)} introduces beneficial noise, and finally, \textbf{Cyclic Training (CT)} promotes diversity across view-specific branches, achieving the best overall results. Each component contributes progressively to the final model's superior performance.


\section{Conclusion}
\label{sec:conclusion}
This work advances Universal Anomaly Detection (Univ-AD) by rethinking the adaptation of state-of-the-art deep learning frameworks for this downstream task, which innovatively makes use of residual features to enable anomaly detection. We introduce self-referential patch-matching, a dynamic process that generates ``intra-residuals'' while simultaneously pruning redundant computations, achieving both higher discrimination power and efficiency. Furthermore, the proposed Self-Navigated Mamba Blocks tailor scanning paths to anomaly contexts, and the multi-view decoder consolidates these features for robust predictions. The proposed algorithm, Self-Navigated Residual Mamba (SNARM), sets new performance records across MVTec AD, MVTec 3D, and VisA datasets against metrics (including Image-AUROC, Pixel-AUROC, PRO, AP) while maintaining real-world deployability. By unifying accuracy and practicality, SNARM addresses a fundamental industrial need: scalable, category-agnostic anomaly detection under resource constraints.

\bibliography{anomaly_detection,interactive_segmentation,others}
\clearpage
\appendix
\section{Appendix}
\label{sec:appendix}


\subsection{Per-Class Quantitative Results in the Multi-Class Setting}
\label{sec:multi_class_result}

We report the per-class performance of our method on the MVTec-AD~\cite{bergmann2019mvtec}, MVTec-3D~\cite{bergmann2021mvtec}, and VisA~\cite{zou_spot--difference_2022} datasets under the multi-class anomaly detection setting. Results of the comparison methods are drawn from MambaAD~\cite{he2024mambaad} and INP-Former~\cite{luo2025exploring}. Table~\ref{tab:mvtec_ad_multi_class} summarizes the image-level anomaly detection and pixel-level localization performance on MVTec-AD, while Table~\ref{tab:mvtec_3d_multi_class} and Table~\ref{tab:visa_multi_class} present the corresponding results on MVTec-3D and VisA. These results highlight the consistent superiority of our approach across diverse object categories.

\subsection{Per-Class Quantitative Results in the Cross-Class Setting}
\label{sec:cross_class_result}

We further evaluate per-class performance in the cross-class anomaly detection setting using the MVTec-AD, MVTec-3D, and VisA datasets. Table~\ref{tab:mvtec_ad_cross_class} reports image-level and pixel-level performance on MVTec-AD, while Table~\ref{tab:mvtec_3d_cross_class} and Table~\ref{tab:visa_cross_class} show results for MVTec-3D and VisA, respectively. Our method consistently outperforms existing approaches, demonstrating strong generalization to unseen categories.

\subsection{Per-Class Quantitative Results in the Single-Class Setting}
\label{sec:single_class_result}

To support future research and reproducibility, we provide detailed per-class performance of SNARM in the single-class setting on MVTec-AD, MVTec-3D, and VisA. Results are shown in Table~\ref{tab:mvtec_ad_single_class}, Table~\ref{tab:mvtec_3d_single_class}, and Table~\ref{tab:visa_single_class}, respectively.

\subsection{Qualitative Results}
\label{sec:qualitative_result}

\begin{figure*} [!tbhp]
\centering
  \includegraphics [width=0.98\textwidth]{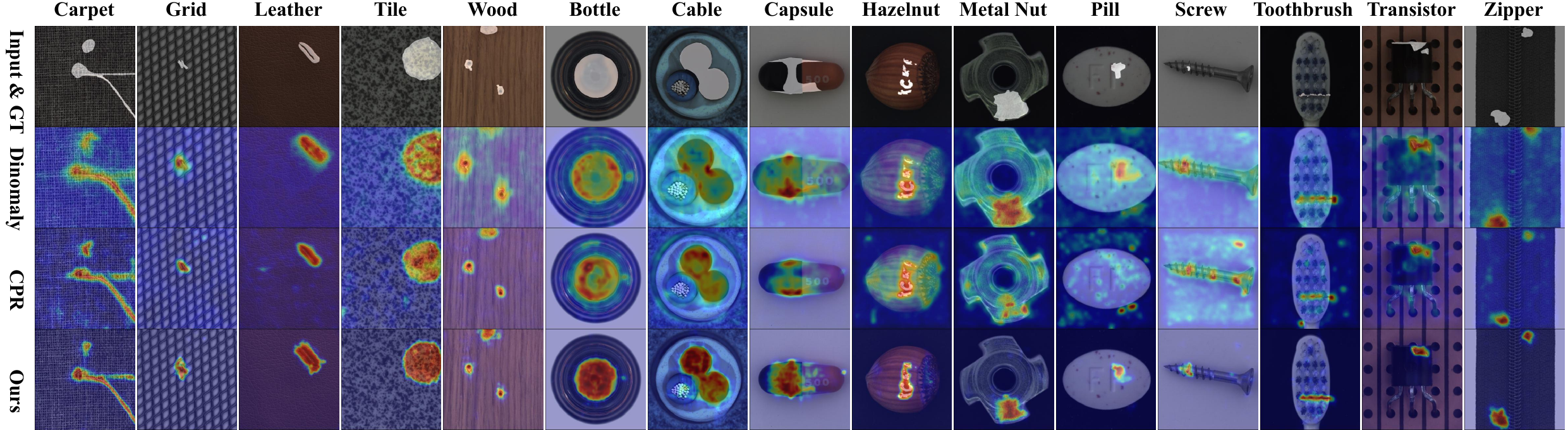}

\caption{
  Qualitative results of our SNARM on MVTec-AD under the Multi-class setting. Two SOTA methods (Dinomaly
  \cite{guo2024dinomaly} and CPR
  \cite{li2023target}) are also involved in the comparison.
}
\vspace*{-0.3cm}
\label{fig:qualitative}
\end{figure*}

Figure~\ref{fig:qualitative} visualizes the anomaly maps generated by our method and the baselines on the MVTec-AD dataset under the multi-class setting. As shown, our method accurately localizes anomalous regions across a wide variety of object categories, demonstrating robust localization capability.

\begin{table}[htbp]
\centering
\begin{tabular}{ccccc} \toprule
Category   & P-AP & PRO  & P-AUROC & I-AUROC \\ \midrule
carpet     & 88.5 & 98.9 & 99.7    & 99.9    \\
grid       & 59.3 & 97.9 & 99.5    & 99.8    \\
leather    & 67.3 & 98.5 & 99.6    & 100.   \\
tile       & 96.0 & 97.7 & 99.6    & 100.   \\
wood       & 82.6 & 96.1 & 98.1    & 96.8    \\
bottle     & 94.8 & 98.5 & 99.5    & 100.   \\
cable      & 86.3 & 96.1 & 99.3    & 99.5    \\
capsule    & 60.8 & 96.5 & 99.0    & 97.3    \\
hazelnut   & 92.1 & 97.4 & 99.7    & 100.   \\
metal\_nut & 91.2 & 96.4 & 98.8    & 99.9    \\
pill       & 83.2 & 98.3 & 98.7    & 98.4    \\
screw      & 54.7 & 96.4 & 99.3    & 98.0    \\
toothbrush & 79.4 & 95.2 & 99.5    & 100.   \\
transistor & 71.1 & 87.9 & 95.0    & 99.3    \\
zipper     & 80.8 & 97.6 & 99.3    & 100.   \\ \midrule
Average    & 79.2 & 96.6 & 99.0    & 99.3   \\ \bottomrule
\end{tabular}
\caption{Per-Class Results on the  MVTec-AD~\cite{bergmann2019mvtec} Dataset under the Single-class setting.}
\label{tab:mvtec_ad_single_class}
\end{table}

\begin{table}[htbp]
\centering
\begin{tabular}{ccccc} \toprule
Category   & P-AP & PRO  & P-AUROC & I-AUROC \\ \midrule
bagel        & 73.5 & 98.1 & 99.5         & 97.2         \\
cable\_gland & 66.0 & 99.3 & 99.8         & 97.5         \\
carrot       & 63.0 & 99.2 & 99.8         & 96.9         \\
cookie       & 71.1 & 95.3 & 98.8         & 89.2         \\
dowel        & 64.7 & 98.2 & 99.6         & 98.7         \\
foam         & 51.2 & 87.1 & 95.4         & 89.6         \\
peach        & 73.8 & 99.2 & 99.8         & 95.4         \\
potato       & 50.8 & 98.4 & 99.6         & 80.8         \\
rope         & 78.9 & 99.2 & 99.8         & 99.5         \\
tire         & 61.4 & 97.8 & 99.6         & 95.1         \\ \midrule
Average      & 65.5 & 97.2 & 99.2         & 94.0        \\ \bottomrule
\end{tabular}
\caption{Per-Class Results on the  MVTec-3D~\cite{bergmann2021mvtec} Dataset under the Single-class setting.}
\label{tab:mvtec_3d_single_class}
\end{table}

\begin{table}[htbp]
\centering
\begin{tabular}{ccccc} \toprule
Category   & P-AP & PRO  & P-AUROC & I-AUROC \\ \midrule
candle      & 54.3 & 97.4 & 99.7         & 99.0         \\
capsules    & 67.7 & 97.3 & 99.7         & 99.0         \\
cashew      & 77.6 & 90.2 & 99.2         & 98.6         \\
chewinggum  & 64.4 & 89.0 & 99.3         & 99.2         \\
fryum       & 42.5 & 88.2 & 95.2         & 98.4         \\
macaroni1   & 31.7 & 97.4 & 99.8         & 98.0         \\
macaroni2   & 20.6 & 96.3 & 99.6         & 90.4         \\
pcb1        & 85.1 & 92.2 & 99.6         & 98.5         \\
pcb2        & 40.6 & 91.8 & 98.8         & 97.6         \\
pcb3        & 52.3 & 84.2 & 99.2         & 99.5         \\
pcb4        & 48.8 & 89.5 & 98.5         & 98.9         \\
pipe\_fryum & 79.6 & 94.4 & 99.6         & 99.1         \\ \midrule
Average     & 55.4 & 92.3 & 99.0         & 98.0         \\ \bottomrule
\end{tabular}
\caption{Per-Class Results on the  VisA~\cite{zou_spot--difference_2022} Dataset under the Single-class setting.}
\label{tab:visa_single_class}
\end{table}

\begin{table*}[!hbt]
\centering
\fontsize{9}{10}\selectfont{
\setlength{\tabcolsep}{0.5mm}{
\begin{tabular}{c|cccc|cc}
\toprule
Method$\rightarrow$                                                                                                                                                                 & DiAD                                                                               & MambaAD                                                                            & Dinomaly                                                                                                                                                                                                     & INP-Former                                                                                                                                                                                                      & \textbf{SNARM} & \textbf{SNARM}\\
Category$\downarrow$ &                                                                                                    \cite{he2024diffusion}                                            & \cite{he2024mambaad}                                              & \cite{guo2024dinomaly}                                                                                                                                                                      & \cite{luo2025exploring}                                                                                                                                                                        & ($T=10^4$)   &  ($T=10^5$)       \\ \hline
carpet                     &42.2/90.6/98.6/99.4&60.0/96.7/99.2/99.8&68.7/97.6/99.3/99.8&72.5/97.7/{\color{blue}{\textbf{99.4}}}/99.9&{\color{blue}{\textbf{85.5}}}/{\color{blue}{\textbf{98.3}}}/{\color{blue}{\textbf{99.4}}}/{\color{red}{\textbf{100.}}}&{\color{red}{\textbf{87.5}}}/{\color{red}{\textbf{98.8}}}/{\color{red}{\textbf{99.6}}}/{\color{red}{\textbf{100.}}}\\
grid                       &{\color{red}{\textbf{66.0}}}/94.0/96.6/98.5&47.4/97.0/99.2/{\color{red}{\textbf{100.}}}&55.3/97.2/99.4/99.9&58.1/{\color{blue}{\textbf{97.7}}}/99.4/99.9&56.4/{\color{blue}{\textbf{97.7}}}/{\color{red}{\textbf{99.5}}}/{\color{red}{\textbf{100.}}}&{\color{blue}{\textbf{59.3}}}/{\color{red}{\textbf{98.1}}}/{\color{red}{\textbf{99.5}}}/{\color{red}{\textbf{100.}}}\\
leather                    &56.1/91.3/98.8/99.8&50.3/{\color{red}{\textbf{98.7}}}/99.4/{\color{red}{\textbf{100.}}}&52.2/97.6/99.4/{\color{red}{\textbf{100.}}}&56.3/98.0/99.4/{\color{red}{\textbf{100.}}}&{\color{blue}{\textbf{62.5}}}/98.5/{\color{blue}{\textbf{99.6}}}/{\color{red}{\textbf{100.}}}&{\color{red}{\textbf{70.6}}}/{\color{red}{\textbf{98.7}}}/{\color{red}{\textbf{99.7}}}/{\color{red}{\textbf{100.}}}\\
tile                       &65.7/90.7/92.4/96.8&45.1/80.0/93.8/98.2&80.1/90.5/98.1/{\color{red}{\textbf{100.}}}&76.6/88.3/97.8/{\color{red}{\textbf{100.}}}&{\color{red}{\textbf{96.2}}}/{\color{blue}{\textbf{97.7}}}/{\color{red}{\textbf{99.6}}}/99.9&{\color{blue}{\textbf{95.6}}}/{\color{red}{\textbf{97.9}}}/{\color{blue}{\textbf{99.5}}}/99.9\\
wood                       &43.3/{\color{red}{\textbf{97.5}}}/93.3/99.7&46.2/91.2/94.4/98.8&72.8/94.0/{\color{blue}{\textbf{97.6}}}/{\color{blue}{\textbf{99.8}}}&74.6/93.7/{\color{blue}{\textbf{97.6}}}/{\color{red}{\textbf{99.9}}}&{\color{blue}{\textbf{74.9}}}/93.4/97.2/98.9&{\color{red}{\textbf{78.6}}}/{\color{blue}{\textbf{95.0}}}/{\color{red}{\textbf{98.3}}}/98.5\\
bottle                     &52.2/86.6/98.4/99.7&79.7/95.2/98.8/{\color{red}{\textbf{100.}}}&88.6/96.6/99.2/{\color{red}{\textbf{100.}}}&88.7/{\color{blue}{\textbf{97.1}}}/99.1/{\color{red}{\textbf{100.}}}&{\color{blue}{\textbf{92.9}}}/96.6/{\color{blue}{\textbf{99.3}}}/{\color{red}{\textbf{100.}}}&{\color{red}{\textbf{93.6}}}/{\color{red}{\textbf{97.6}}}/{\color{red}{\textbf{99.4}}}/{\color{red}{\textbf{100.}}}\\
cable                      &50.1/80.5/96.8/94.8&42.2/90.3/95.8/98.8&72.0/94.2/98.6/{\color{red}{\textbf{100.}}}&{\color{blue}{\textbf{79.3}}}/{\color{blue}{\textbf{94.4}}}/{\color{blue}{\textbf{98.8}}}/{\color{red}{\textbf{100.}}}&78.7/94.2/98.5/99.6&{\color{red}{\textbf{82.2}}}/{\color{red}{\textbf{95.9}}}/{\color{red}{\textbf{99.1}}}/99.3\\
capsule                    &42.0/87.2/97.1/89.0&43.9/92.6/98.4/94.4&61.4/97.2/98.7/{\color{blue}{\textbf{97.9}}}&60.3/{\color{red}{\textbf{97.7}}}/{\color{blue}{\textbf{98.8}}}/{\color{red}{\textbf{99.0}}}&{\color{blue}{\textbf{62.5}}}/95.3/{\color{blue}{\textbf{98.8}}}/94.9&{\color{red}{\textbf{65.9}}}/{\color{blue}{\textbf{97.4}}}/{\color{red}{\textbf{99.3}}}/96.5\\
hazelnut                   &79.2/91.5/98.3/99.5&63.6/95.7/99.0/{\color{red}{\textbf{100.}}}&82.2/{\color{blue}{\textbf{97.0}}}/99.4/{\color{red}{\textbf{100.}}}&81.8/{\color{blue}{\textbf{97.0}}}/99.5/{\color{red}{\textbf{100.}}}&{\color{blue}{\textbf{91.3}}}/95.5/{\color{blue}{\textbf{99.6}}}/99.9&{\color{red}{\textbf{92.1}}}/{\color{red}{\textbf{97.7}}}/{\color{red}{\textbf{99.7}}}/99.7\\
metal Nut                  &30.0/90.6/97.3/99.1&74.5/93.7/96.7/99.9&78.6/94.9/96.9/{\color{red}{\textbf{100.}}}&81.2/95.1/97.5/{\color{red}{\textbf{100.}}}&{\color{red}{\textbf{93.5}}}/{\color{red}{\textbf{96.8}}}/{\color{blue}{\textbf{99.0}}}/{\color{red}{\textbf{100.}}}&{\color{blue}{\textbf{91.8}}}/{\color{blue}{\textbf{96.6}}}/{\color{red}{\textbf{99.1}}}/{\color{red}{\textbf{100.}}}\\
pill                       &46.0/89.0/95.7/95.7&64.0/95.7/97.4/97.0&76.4/{\color{blue}{\textbf{97.3}}}/97.8/99.1&76.1/{\color{blue}{\textbf{97.3}}}/97.7/99.1&{\color{red}{\textbf{82.1}}}/96.2/{\color{red}{\textbf{98.6}}}/{\color{blue}{\textbf{99.7}}}&{\color{blue}{\textbf{80.1}}}/{\color{red}{\textbf{98.1}}}/{\color{blue}{\textbf{98.5}}}/{\color{red}{\textbf{99.8}}}\\
screw                      &{\color{blue}{\textbf{60.6}}}/95.0/97.9/90.7&49.8/97.1/{\color{blue}{\textbf{99.5}}}/94.7&60.2/{\color{red}{\textbf{98.3}}}/{\color{red}{\textbf{99.6}}}/{\color{red}{\textbf{98.4}}}&{\color{red}{\textbf{61.8}}}/{\color{blue}{\textbf{97.9}}}/{\color{blue}{\textbf{99.5}}}/97.5&48.4/93.1/98.7/93.7&58.6/96.4/99.4/{\color{blue}{\textbf{97.7}}}\\
toothbrush                 &78.7/95.0/99.0/99.7&48.5/91.7/99.0/98.3&51.5/{\color{blue}{\textbf{95.3}}}/98.9/{\color{red}{\textbf{100.}}}&58.3/{\color{red}{\textbf{95.9}}}/99.1/{\color{red}{\textbf{100.}}}&{\color{red}{\textbf{79.2}}}/92.0/{\color{blue}{\textbf{99.4}}}/{\color{red}{\textbf{100.}}}&{\color{red}{\textbf{79.2}}}/94.3/{\color{red}{\textbf{99.5}}}/{\color{red}{\textbf{100.}}}\\
transistor                 &15.6/{\color{red}{\textbf{90.0}}}/95.1/{\color{blue}{\textbf{99.8}}}&69.4/87.0/96.5/{\color{red}{\textbf{100.}}}&59.9/77.0/93.2/99.0&64.0/79.0/94.7/99.7&{\color{red}{\textbf{74.5}}}/88.1/{\color{blue}{\textbf{96.6}}}/{\color{blue}{\textbf{99.8}}}&{\color{blue}{\textbf{72.3}}}/{\color{blue}{\textbf{89.8}}}/{\color{red}{\textbf{97.0}}}/99.6\\
zipper                     &60.7/91.6/96.2/95.1&60.4/94.3/98.4/99.3&{\color{red}{\textbf{79.5}}}/{\color{blue}{\textbf{97.2}}}/{\color{red}{\textbf{99.2}}}/{\color{red}{\textbf{100.}}}&75.8/96.4/99.0/{\color{red}{\textbf{100.}}}&75.2/95.4/98.7/{\color{red}{\textbf{100.}}}&{\color{blue}{\textbf{78.0}}}/{\color{red}{\textbf{97.4}}}/{\color{red}{\textbf{99.2}}}/{\color{red}{\textbf{100.}}}\\ \midrule
Average                       &52.6/90.7/96.8/97.2&56.3/93.1/97.7/98.6&69.3/94.8/98.4/{\color{red}{\textbf{99.6}}}&71.0/94.9/98.5/{\color{red}{\textbf{99.6}}}&{\color{blue}{\textbf{76.9}}}/{\color{blue}{\textbf{95.3}}}/{\color{blue}{\textbf{98.8}}}/99.1&{\color{red}{\textbf{79.0}}}/{\color{red}{\textbf{96.6}}}/{\color{red}{\textbf{99.1}}}/99.4\\
\bottomrule
\end{tabular}}}
\caption{Per-Class Results on the  MVTec-AD~\cite{bergmann2019mvtec} Dataset under the Multi-class setting with P-AP/PRO/P-AUROC/I-AUROC metrics.}
\label{tab:mvtec_ad_multi_class}
\end{table*}

\begin{table*}[!ht]
\centering
\fontsize{9}{10}\selectfont{
\setlength{\tabcolsep}{0.5mm}{
\begin{tabular}{c|cccc|cc}
\toprule
Method$\rightarrow$                                                                                                                                                                 & DiAD                                                                               & MambaAD                                                                            & Dinomaly                                                                                                                                                                                                     & INP-Former                                                                                                                                                                                                      & \textbf{SNARM} & \textbf{SNARM}\\
Category$\downarrow$ &                                                                                                    \cite{he2024diffusion}                                            & \cite{he2024mambaad}                                              & \cite{guo2024dinomaly}                                                                                                                                                                      & \cite{luo2025exploring}                                                                                                                                                                        & ($T=10^4$)   &  ($T=10^5$)       \\ \hline
bagel        &49.6/93.8/98.5/{\color{red}{\textbf{100.}}}&38.3/92.1/98.5/87.7&59.1/95.9/{\color{red}{\textbf{99.3}}}/90.2&59.5/{\color{blue}{\textbf{96.9}}}/{\color{red}{\textbf{99.3}}}/92.6&{\color{blue}{\textbf{66.4}}}/96.1/99.2/91.5&{\color{red}{\textbf{69.1}}}/{\color{red}{\textbf{97.7}}}/{\color{red}{\textbf{99.3}}}/{\color{blue}{\textbf{95.4}}}\\
cable\_gland &25.2/94.5/98.4/68.1&39.5/98.4/99.4/94.3&62.5/99.3/{\color{red}{\textbf{99.8}}}/98.1&60.9/{\color{red}{\textbf{99.4}}}/{\color{red}{\textbf{99.8}}}/{\color{red}{\textbf{98.6}}}&{\color{blue}{\textbf{62.6}}}/99.1/99.7/96.1&{\color{red}{\textbf{67.9}}}/{\color{red}{\textbf{99.4}}}/{\color{red}{\textbf{99.8}}}/{\color{red}{\textbf{98.6}}}\\
carrot       &20.0/94.6/98.6/{\color{blue}{\textbf{94.4}}}&30.1/98.1/99.4/90.7&44.7/98.4/{\color{blue}{\textbf{99.7}}}/91.8&47.0/98.9/{\color{blue}{\textbf{99.7}}}/94.3&{\color{blue}{\textbf{56.1}}}/{\color{blue}{\textbf{99.0}}}/{\color{blue}{\textbf{99.7}}}/94.2&{\color{red}{\textbf{60.6}}}/{\color{red}{\textbf{99.2}}}/{\color{red}{\textbf{99.8}}}/{\color{red}{\textbf{97.8}}}\\
cookie       &14.0/83.5/94.3/69.4&39.0/83.6/96.8/61.2&58.8/91.4/98.1/78.2&64.6/92.2/{\color{blue}{\textbf{98.3}}}/79.9&{\color{red}{\textbf{72.2}}}/{\color{red}{\textbf{95.3}}}/{\color{red}{\textbf{98.4}}}/{\color{blue}{\textbf{87.3}}}&{\color{blue}{\textbf{71.7}}}/{\color{red}{\textbf{95.3}}}/{\color{blue}{\textbf{98.3}}}/{\color{red}{\textbf{90.9}}}\\
dowel        &31.4/89.6/97.2/98.0&49.9/97.1/99.6/97.6&{\color{red}{\textbf{66.1}}}/{\color{red}{\textbf{98.9}}}/{\color{red}{\textbf{99.7}}}/{\color{blue}{\textbf{99.1}}}&63.0/{\color{red}{\textbf{98.9}}}/{\color{red}{\textbf{99.7}}}/99.0&59.3/97.4/99.4/94.0&{\color{blue}{\textbf{65.2}}}/{\color{red}{\textbf{98.9}}}/{\color{red}{\textbf{99.7}}}/{\color{red}{\textbf{99.2}}}\\
foam         &9.6/69.1/89.8/{\color{red}{\textbf{100.}}}&23.4/82.7/95.1/84.0&41.8/{\color{blue}{\textbf{88.8}}}/{\color{red}{\textbf{96.6}}}/89.1&41.0/{\color{red}{\textbf{88.9}}}/96.3/89.4&{\color{blue}{\textbf{44.4}}}/86.7/95.6/90.2&{\color{red}{\textbf{47.4}}}/88.7/{\color{blue}{\textbf{96.4}}}/{\color{blue}{\textbf{90.3}}}\\
peach        &27.6/94.2/98.4/58.0&43.2/97.1/99.4/92.8&56.1/98.1/99.6/82.7&63.9/{\color{blue}{\textbf{99.0}}}/{\color{red}{\textbf{99.8}}}/{\color{red}{\textbf{96.3}}}&{\color{blue}{\textbf{68.6}}}/98.8/99.7/89.0&{\color{red}{\textbf{75.6}}}/{\color{red}{\textbf{99.3}}}/{\color{red}{\textbf{99.8}}}/{\color{blue}{\textbf{95.5}}}\\
potato       &8.6/93.9/98.0/76.3&17.6/94.8/99.0/66.8&44.5/97.7/99.5/{\color{red}{\textbf{80.1}}}&46.4/{\color{blue}{\textbf{98.2}}}/{\color{red}{\textbf{99.7}}}/{\color{red}{\textbf{80.1}}}&{\color{blue}{\textbf{46.6}}}/97.6/99.5/70.4&{\color{red}{\textbf{50.6}}}/{\color{red}{\textbf{98.4}}}/{\color{blue}{\textbf{99.6}}}/79.5\\
rope         &61.0/96.5/99.3/89.2&52.1/95.5/99.4/97.4&58.4/97.9/99.5/{\color{blue}{\textbf{99.3}}}&56.6/{\color{blue}{\textbf{98.7}}}/99.5/{\color{red}{\textbf{99.5}}}&{\color{blue}{\textbf{71.2}}}/98.6/{\color{blue}{\textbf{99.6}}}/98.7&{\color{red}{\textbf{73.8}}}/{\color{red}{\textbf{98.9}}}/{\color{red}{\textbf{99.7}}}/99.2\\
tire         &5.9/68.8/91.8/92.7&42.0/97.0/99.5/90.0&{\color{red}{\textbf{58.1}}}/{\color{red}{\textbf{99.0}}}/{\color{red}{\textbf{99.8}}}/{\color{red}{\textbf{97.3}}}&{\color{blue}{\textbf{56.3}}}/{\color{red}{\textbf{99.0}}}/{\color{red}{\textbf{99.8}}}/{\color{blue}{\textbf{96.5}}}&47.6/96.9/99.4/90.4&54.5/98.0/99.6/92.6\\ \midrule
Average         &25.3/87.8/96.4/84.6&37.5/93.6/98.6/86.2&55.0/96.5/{\color{red}{\textbf{99.2}}}/90.6&55.9/{\color{blue}{\textbf{97.0}}}/{\color{red}{\textbf{99.2}}}/{\color{blue}{\textbf{92.6}}}&{\color{blue}{\textbf{59.5}}}/96.5/99.0/90.2&{\color{red}{\textbf{63.6}}}/{\color{red}{\textbf{97.4}}}/{\color{red}{\textbf{99.2}}}/{\color{red}{\textbf{93.9}}}\\
\bottomrule
\end{tabular}}}
\caption{Per-Class Results on the  MVTec-3D~\cite{bergmann2021mvtec} Dataset under the Multi-class setting with P-AP/PRO/P-AUROC/I-AUROC metrics.}
\label{tab:mvtec_3d_multi_class}
\end{table*}
\begin{table*}[!ht]
\centering
\fontsize{9}{10}\selectfont{
\setlength{\tabcolsep}{0.5mm}{
\begin{tabular}{c|cccc|cc}
\toprule
Method$\rightarrow$                                                                                                                                                                 & DiAD                                                                               & MambaAD                                                                            & Dinomaly                                                                                                                                                                                                     & INP-Former                                                                                                                                                                                                      & \textbf{SNARM} & \textbf{SNARM}\\
Category$\downarrow$ &                                                                                                    \cite{he2024diffusion}                                            & \cite{he2024mambaad}                                              & \cite{guo2024dinomaly}                                                                                                                                                                      & \cite{luo2025exploring}                                                                                                                                                                        & ($T=10^4$)   &  ($T=10^5$)       \\ \hline
candle                      &12.8/89.4/97.3/92.8&23.2/95.5/99.0/96.8&43.0/95.4/{\color{blue}{\textbf{99.4}}}/{\color{red}{\textbf{98.7}}}&43.9/95.6/{\color{blue}{\textbf{99.4}}}/98.4&{\color{red}{\textbf{53.5}}}/{\color{blue}{\textbf{95.9}}}/99.2/98.0&{\color{blue}{\textbf{52.4}}}/{\color{red}{\textbf{96.2}}}/{\color{red}{\textbf{99.7}}}/{\color{blue}{\textbf{98.6}}}\\
capsules                    &10.0/77.9/97.3/58.2&61.3/91.8/99.1/91.8&65.0/97.4/{\color{red}{\textbf{99.6}}}/{\color{blue}{\textbf{98.6}}}&{\color{blue}{\textbf{67.2}}}/{\color{blue}{\textbf{98.0}}}/{\color{red}{\textbf{99.6}}}/{\color{red}{\textbf{99.1}}}&{\color{red}{\textbf{68.3}}}/96.7/{\color{red}{\textbf{99.6}}}/97.9&66.8/{\color{red}{\textbf{98.1}}}/{\color{red}{\textbf{99.6}}}/98.2\\
cashew                      &53.1/61.8/90.9/91.5&46.8/87.8/94.3/94.5&64.5/{\color{blue}{\textbf{94.0}}}/97.1/{\color{red}{\textbf{98.7}}}&66.2/92.0/97.7/{\color{blue}{\textbf{98.6}}}&{\color{red}{\textbf{73.1}}}/85.2/{\color{red}{\textbf{98.9}}}/96.4&{\color{blue}{\textbf{68.2}}}/{\color{red}{\textbf{94.6}}}/{\color{blue}{\textbf{98.8}}}/{\color{blue}{\textbf{98.6}}}\\
chewinggum                  &11.9/59.5/94.7/99.1&57.5/79.7/98.1/97.7&65.0/88.1/99.1/{\color{red}{\textbf{99.8}}}&59.6/86.5/98.9/{\color{blue}{\textbf{99.7}}}&{\color{blue}{\textbf{86.0}}}/{\color{blue}{\textbf{89.3}}}/{\color{blue}{\textbf{99.3}}}/99.3&{\color{red}{\textbf{86.4}}}/{\color{red}{\textbf{92.6}}}/{\color{red}{\textbf{99.5}}}/99.2\\
fryum                       &{\color{red}{\textbf{58.6}}}/81.3/{\color{red}{\textbf{97.6}}}/89.8&47.8/91.6/{\color{blue}{\textbf{96.9}}}/95.2&{\color{blue}{\textbf{51.6}}}/{\color{blue}{\textbf{93.5}}}/96.6/{\color{blue}{\textbf{98.8}}}&51.2/{\color{red}{\textbf{94.2}}}/96.8/{\color{red}{\textbf{99.3}}}&47.1/85.6/{\color{blue}{\textbf{96.9}}}/98.2&46.6/89.0/96.7/98.5\\
macaron1                    &10.2/68.5/94.1/85.7&17.5/95.2/99.5/91.6&{\color{blue}{\textbf{33.5}}}/{\color{blue}{\textbf{96.4}}}/{\color{red}{\textbf{99.6}}}/{\color{blue}{\textbf{98.0}}}&{\color{red}{\textbf{33.9}}}/96.0/{\color{red}{\textbf{99.6}}}/{\color{red}{\textbf{98.5}}}&24.0/93.6/98.9/95.8&20.1/{\color{red}{\textbf{97.0}}}/{\color{red}{\textbf{99.6}}}/95.1\\
macaron2                    &0.9/73.1/93.6/62.5&9.2/96.2/99.5/81.6&{\color{blue}{\textbf{24.7}}}/{\color{red}{\textbf{98.7}}}/{\color{blue}{\textbf{99.7}}}/{\color{blue}{\textbf{95.9}}}&{\color{red}{\textbf{26.8}}}/{\color{red}{\textbf{98.7}}}/{\color{red}{\textbf{99.8}}}/{\color{red}{\textbf{96.9}}}&15.9/91.8/95.9/90.5&16.5/97.5/99.4/92.0\\
pcb1                        &49.6/80.2/98.7/88.1&77.1/92.8/{\color{red}{\textbf{99.8}}}/95.4&87.9/{\color{blue}{\textbf{95.1}}}/99.5/{\color{red}{\textbf{99.1}}}&87.6/{\color{red}{\textbf{95.2}}}/99.6/{\color{blue}{\textbf{98.8}}}&{\color{red}{\textbf{88.7}}}/87.1/99.2/96.0&{\color{blue}{\textbf{88.2}}}/94.5/{\color{blue}{\textbf{99.7}}}/98.0\\
pcb2                        &7.5/67.0/95.2/91.4&13.3/89.6/{\color{red}{\textbf{98.9}}}/94.2&{\color{red}{\textbf{47.0}}}/{\color{blue}{\textbf{91.3}}}/98.0/{\color{red}{\textbf{99.3}}}&31.2/{\color{red}{\textbf{91.9}}}/{\color{blue}{\textbf{98.7}}}/{\color{blue}{\textbf{98.8}}}&{\color{blue}{\textbf{44.0}}}/87.8/97.8/96.1&42.8/89.8/98.6/96.1\\
pcb3                        &8.0/68.9/96.7/86.2&18.3/89.1/{\color{blue}{\textbf{99.1}}}/93.7&41.7/{\color{red}{\textbf{94.6}}}/98.4/{\color{blue}{\textbf{98.9}}}&30.6/{\color{blue}{\textbf{94.3}}}/98.8/{\color{red}{\textbf{99.2}}}&{\color{blue}{\textbf{46.9}}}/86.0/98.1/97.1&{\color{red}{\textbf{52.9}}}/90.5/{\color{red}{\textbf{99.2}}}/{\color{blue}{\textbf{98.9}}}\\
pcb4                        &17.6/85.0/97.0/99.6&47.0/87.6/98.6/{\color{red}{\textbf{99.9}}}&{\color{blue}{\textbf{50.5}}}/{\color{red}{\textbf{94.4}}}/{\color{blue}{\textbf{98.7}}}/99.8&{\color{red}{\textbf{53.2}}}/{\color{blue}{\textbf{94.2}}}/{\color{red}{\textbf{98.8}}}/{\color{red}{\textbf{99.9}}}&42.2/87.4/97.7/98.8&49.3/91.0/{\color{blue}{\textbf{98.7}}}/99.1\\
pipe\_fryum                 &72.7/89.9/99.4/96.2&53.5/95.1/99.1/98.7&64.3/95.2/99.2/{\color{blue}{\textbf{99.2}}}&63.3/{\color{blue}{\textbf{95.8}}}/99.3/{\color{red}{\textbf{99.5}}}&{\color{red}{\textbf{77.9}}}/94.6/{\color{red}{\textbf{99.6}}}/97.1&{\color{blue}{\textbf{75.5}}}/{\color{red}{\textbf{96.5}}}/{\color{red}{\textbf{99.6}}}/99.0\\ \midrule
Average &26.1/75.2/96.0/86.8&39.4/91.0/98.5/94.3&53.2/{\color{red}{\textbf{94.5}}}/98.7/{\color{blue}{\textbf{98.7}}}&51.2/{\color{blue}{\textbf{94.4}}}/{\color{blue}{\textbf{98.9}}}/{\color{red}{\textbf{98.9}}}&{\color{red}{\textbf{55.6}}}/90.1/98.4/96.8&{\color{blue}{\textbf{55.5}}}/94.0/{\color{red}{\textbf{99.1}}}/97.6\\

\bottomrule
\end{tabular}}}
\caption{Per-Class Results on the  VisA~\cite{zou_spot--difference_2022} Dataset under the Multi-class setting with P-AP/PRO/P-AUROC/I-AUROC metrics.}
\label{tab:visa_multi_class}
\end{table*}


\begin{table*}[!ht]
\centering
\fontsize{9}{10}\selectfont{
\begin{tabular}{c|ccc|c}
\toprule
Method$\rightarrow$      & CPR                 & HETMM               & Dinomaly            & \textbf{SNARM}              \\
Category$\downarrow$ & \cite{li2023target} & \cite{chen2024hard}&\cite{guo2024dinomaly}&($T=10^4$)\\\bottomrule
carpet     &65.1/92.4/97.1/92.9&{\color{blue}{\textbf{67.2}}}/{\color{blue}{\textbf{97.2}}}/{\color{blue}{\textbf{99.1}}}/{\color{blue}{\textbf{99.9}}}&45.2/84.9/95.8/98.0&{\color{red}{\textbf{87.1}}}/{\color{red}{\textbf{98.7}}}/{\color{red}{\textbf{99.7}}}/{\color{red}{\textbf{100.}}}\\
grid       &15.9/74.6/92.1/74.6&{\color{blue}{\textbf{48.0}}}/{\color{blue}{\textbf{96.6}}}/{\color{blue}{\textbf{98.8}}}/{\color{blue}{\textbf{96.9}}}&1.6/13.6/47.5/63.8&{\color{red}{\textbf{57.2}}}/{\color{red}{\textbf{98.0}}}/{\color{red}{\textbf{99.5}}}/{\color{red}{\textbf{99.8}}}\\
leather    &66.9/{\color{red}{\textbf{99.4}}}/99.6/{\color{red}{\textbf{100.}}}&{\color{blue}{\textbf{67.2}}}/98.7/{\color{blue}{\textbf{99.7}}}/{\color{red}{\textbf{100.}}}&10.7/34.3/73.7/{\color{red}{\textbf{100.}}}&{\color{red}{\textbf{78.7}}}/{\color{blue}{\textbf{99.0}}}/{\color{red}{\textbf{99.8}}}/{\color{red}{\textbf{100.}}}\\
tile       &46.6/79.5/89.7/93.4&{\color{blue}{\textbf{70.3}}}/{\color{blue}{\textbf{90.6}}}/{\color{blue}{\textbf{97.3}}}/{\color{blue}{\textbf{99.7}}}&46.2/58.0/80.6/92.0&{\color{red}{\textbf{95.7}}}/{\color{red}{\textbf{97.6}}}/{\color{red}{\textbf{99.5}}}/{\color{red}{\textbf{100.}}}\\
wood       &{\color{blue}{\textbf{61.7}}}/{\color{red}{\textbf{95.0}}}/95.1/{\color{blue}{\textbf{98.8}}}&58.1/{\color{red}{\textbf{95.0}}}/{\color{blue}{\textbf{96.3}}}/{\color{red}{\textbf{99.7}}}&18.0/23.9/58.8/66.8&{\color{red}{\textbf{78.7}}}/90.9/{\color{red}{\textbf{97.9}}}/96.8\\
bottle     &{\color{blue}{\textbf{89.6}}}/{\color{blue}{\textbf{97.5}}}/{\color{blue}{\textbf{99.1}}}/{\color{red}{\textbf{100.}}}&83.8/96.3/98.8/{\color{red}{\textbf{100.}}}&13.8/30.9/73.2/55.0&{\color{red}{\textbf{94.4}}}/{\color{red}{\textbf{98.0}}}/{\color{red}{\textbf{99.5}}}/{\color{red}{\textbf{100.}}}\\
cable      &{\color{blue}{\textbf{74.3}}}/88.0/98.3/92.3&67.9/{\color{blue}{\textbf{94.3}}}/{\color{blue}{\textbf{98.6}}}/{\color{red}{\textbf{99.7}}}&6.5/23.5/54.2/52.2&{\color{red}{\textbf{85.0}}}/{\color{red}{\textbf{96.1}}}/{\color{red}{\textbf{99.2}}}/{\color{blue}{\textbf{99.6}}}\\
capsule    &46.6/94.3/{\color{blue}{\textbf{98.8}}}/89.0&{\color{blue}{\textbf{50.0}}}/{\color{blue}{\textbf{96.4}}}/{\color{blue}{\textbf{98.8}}}/{\color{red}{\textbf{100.}}}&4.9/46.5/86.5/46.8&{\color{red}{\textbf{62.7}}}/{\color{red}{\textbf{97.4}}}/{\color{red}{\textbf{99.3}}}/{\color{blue}{\textbf{97.6}}}\\
hazelnut   &73.1/95.8/99.1/{\color{blue}{\textbf{99.3}}}&{\color{blue}{\textbf{76.4}}}/{\color{blue}{\textbf{96.4}}}/{\color{blue}{\textbf{99.4}}}/{\color{red}{\textbf{100.}}}&13.5/39.2/84.3/72.4&{\color{red}{\textbf{90.6}}}/{\color{red}{\textbf{96.5}}}/{\color{red}{\textbf{99.6}}}/98.9\\
metal\_nut &83.1/91.4/96.7/96.0&{\color{blue}{\textbf{87.6}}}/{\color{blue}{\textbf{94.0}}}/{\color{blue}{\textbf{98.3}}}/{\color{red}{\textbf{100.}}}&33.3/48.3/76.0/61.4&{\color{red}{\textbf{96.2}}}/{\color{red}{\textbf{95.9}}}/{\color{red}{\textbf{99.4}}}/{\color{red}{\textbf{100.}}}\\
pill       &50.0/89.0/94.6/73.6&{\color{blue}{\textbf{77.0}}}/{\color{blue}{\textbf{97.0}}}/{\color{red}{\textbf{98.4}}}/{\color{blue}{\textbf{98.7}}}&8.8/47.5/74.8/44.2&{\color{red}{\textbf{81.3}}}/{\color{red}{\textbf{98.0}}}/{\color{blue}{\textbf{98.3}}}/{\color{red}{\textbf{100.}}}\\
screw      &35.9/{\color{red}{\textbf{96.8}}}/{\color{red}{\textbf{99.3}}}/88.3&{\color{red}{\textbf{53.1}}}/{\color{blue}{\textbf{96.5}}}/{\color{red}{\textbf{99.3}}}/{\color{blue}{\textbf{93.1}}}&1.4/57.8/87.7/60.4&{\color{blue}{\textbf{51.3}}}/96.3/{\color{red}{\textbf{99.3}}}/{\color{red}{\textbf{97.0}}}\\
toothbrush &60.9/{\color{red}{\textbf{95.5}}}/99.1/{\color{red}{\textbf{100.}}}&{\color{blue}{\textbf{65.7}}}/{\color{blue}{\textbf{94.8}}}/{\color{red}{\textbf{99.4}}}/{\color{red}{\textbf{100.}}}&12.6/40.4/86.8/78.9&{\color{red}{\textbf{72.6}}}/{\color{blue}{\textbf{94.8}}}/{\color{red}{\textbf{99.4}}}/{\color{red}{\textbf{100.}}}\\
transistor &{\color{blue}{\textbf{74.0}}}/87.0/94.5/{\color{blue}{\textbf{99.5}}}&{\color{red}{\textbf{82.1}}}/{\color{red}{\textbf{96.1}}}/{\color{red}{\textbf{99.0}}}/{\color{red}{\textbf{100.}}}&7.0/17.0/52.6/31.4&66.9/{\color{blue}{\textbf{87.1}}}/{\color{blue}{\textbf{95.5}}}/99.3\\
zipper     &{\color{red}{\textbf{78.1}}}/95.1/{\color{blue}{\textbf{98.7}}}/94.8&65.5/{\color{blue}{\textbf{96.6}}}/98.6/{\color{blue}{\textbf{99.7}}}&5.3/27.0/73.7/31.2&{\color{blue}{\textbf{78.0}}}/{\color{red}{\textbf{97.1}}}/{\color{red}{\textbf{99.1}}}/{\color{red}{\textbf{100.}}}\\ \midrule
Average    &61.4/91.4/96.8/92.8&{\color{blue}{\textbf{68.0}}}/{\color{blue}{\textbf{95.8}}}/{\color{blue}{\textbf{98.6}}}/{\color{blue}{\textbf{99.2}}}&15.2/39.5/73.8/63.6&{\color{red}{\textbf{78.4}}}/{\color{red}{\textbf{96.1}}}/{\color{red}{\textbf{99.0}}}/{\color{red}{\textbf{99.3}}}\\
\bottomrule
\end{tabular}}
\caption{Per-Class Results on the  MVTec-AD~\cite{bergmann2019mvtec} Dataset under the Cross-class setting with P-AP/PRO/P-AUROC/I-AUROC metrics.}
\label{tab:mvtec_ad_cross_class}
\end{table*}
\begin{table*}[!ht]
\centering
\fontsize{9}{10}\selectfont{
\begin{tabular}{c|ccc|c}
\toprule
Method$\rightarrow$      & CPR                 & HETMM               & Dinomaly            & \textbf{SNARM}              \\
Category$\downarrow$ & \cite{li2023target} & \cite{chen2024hard}&\cite{guo2024dinomaly}&($T=10^4$)\\\bottomrule
cable\_gland &{\color{blue}{\textbf{64.2}}}/{\color{blue}{\textbf{97.5}}}/{\color{blue}{\textbf{99.0}}}/85.6&52.7/93.2/98.0/{\color{red}{\textbf{94.3}}}&1.1/48.2/85.0/46.4&{\color{red}{\textbf{77.8}}}/{\color{red}{\textbf{98.8}}}/{\color{red}{\textbf{99.6}}}/{\color{blue}{\textbf{93.1}}}\\
bagel        &7.8/89.9/96.4/69.8&{\color{blue}{\textbf{25.8}}}/{\color{blue}{\textbf{96.9}}}/{\color{blue}{\textbf{98.8}}}/{\color{blue}{\textbf{89.0}}}&4.5/50.8/90.0/49.0&{\color{red}{\textbf{75.4}}}/{\color{red}{\textbf{99.7}}}/{\color{red}{\textbf{99.9}}}/{\color{red}{\textbf{97.2}}}\\
carrot       &10.8/95.7/98.7/65.1&{\color{blue}{\textbf{27.9}}}/{\color{blue}{\textbf{97.7}}}/{\color{blue}{\textbf{99.2}}}/{\color{blue}{\textbf{87.1}}}&10.2/88.3/97.5/58.8&{\color{red}{\textbf{63.8}}}/{\color{red}{\textbf{99.3}}}/{\color{red}{\textbf{99.8}}}/{\color{red}{\textbf{95.9}}}\\
foam         &22.9/89.9/96.8/60.6&{\color{blue}{\textbf{51.5}}}/{\color{blue}{\textbf{90.6}}}/{\color{blue}{\textbf{97.7}}}/{\color{blue}{\textbf{65.3}}}&0.7/26.1/73.1/53.8&{\color{red}{\textbf{75.1}}}/{\color{red}{\textbf{95.7}}}/{\color{red}{\textbf{98.8}}}/{\color{red}{\textbf{89.0}}}\\
rope         &21.4/97.0/{\color{blue}{\textbf{99.3}}}/90.2&{\color{blue}{\textbf{49.6}}}/{\color{blue}{\textbf{97.5}}}/99.1/{\color{blue}{\textbf{98.3}}}&2.4/54.2/87.0/21.6&{\color{red}{\textbf{72.9}}}/{\color{red}{\textbf{99.3}}}/{\color{red}{\textbf{99.8}}}/{\color{red}{\textbf{99.4}}}\\
dowel        &{\color{blue}{\textbf{29.2}}}/83.7/{\color{blue}{\textbf{95.3}}}/{\color{blue}{\textbf{78.8}}}&28.5/{\color{red}{\textbf{86.8}}}/{\color{red}{\textbf{95.6}}}/77.8&3.7/81.2/94.9/52.4&{\color{red}{\textbf{50.3}}}/{\color{blue}{\textbf{86.4}}}/94.3/{\color{red}{\textbf{92.6}}}\\
potato       &10.8/92.4/97.6/55.5&{\color{blue}{\textbf{31.9}}}/{\color{blue}{\textbf{94.5}}}/{\color{blue}{\textbf{98.3}}}/{\color{blue}{\textbf{79.9}}}&3.0/83.8/95.8/33.3&{\color{red}{\textbf{81.0}}}/{\color{red}{\textbf{99.6}}}/{\color{red}{\textbf{99.9}}}/{\color{red}{\textbf{97.9}}}\\
cookie       &0.5/53.0/84.7/50.2&{\color{blue}{\textbf{14.8}}}/{\color{blue}{\textbf{96.5}}}/{\color{blue}{\textbf{98.9}}}/{\color{blue}{\textbf{64.9}}}&8.1/72.5/90.6/50.9&{\color{red}{\textbf{55.7}}}/{\color{red}{\textbf{98.4}}}/{\color{red}{\textbf{99.7}}}/{\color{red}{\textbf{84.2}}}\\
tire         &53.5/90.5/98.4/90.4&{\color{blue}{\textbf{60.4}}}/{\color{blue}{\textbf{97.8}}}/{\color{blue}{\textbf{99.6}}}/{\color{blue}{\textbf{98.1}}}&0.4/9.0/68.2/39.4&{\color{red}{\textbf{77.8}}}/{\color{red}{\textbf{98.8}}}/{\color{red}{\textbf{99.8}}}/{\color{red}{\textbf{99.6}}}\\
peach        &14.6/86.1/96.8/67.3&{\color{blue}{\textbf{33.6}}}/{\color{blue}{\textbf{96.9}}}/{\color{blue}{\textbf{99.2}}}/{\color{blue}{\textbf{79.6}}}&7.3/80.9/95.5/62.9&{\color{red}{\textbf{59.4}}}/{\color{red}{\textbf{97.0}}}/{\color{red}{\textbf{99.3}}}/{\color{red}{\textbf{96.4}}}\\ \midrule
Average      &23.6/87.6/96.3/71.3&{\color{blue}{\textbf{37.7}}}/{\color{blue}{\textbf{94.8}}}/{\color{blue}{\textbf{98.4}}}/{\color{blue}{\textbf{83.4}}}&4.2/59.5/87.8/46.9&{\color{red}{\textbf{68.9}}}/{\color{red}{\textbf{97.3}}}/{\color{red}{\textbf{99.1}}}/{\color{red}{\textbf{94.5}}}\\
\bottomrule
\end{tabular}}
\caption{Per-Class Results on the  MVTec-3D~\cite{bergmann2021mvtec} Dataset under the Cross-class setting with P-AP/PRO/P-AUROC/I-AUROC metrics.}
\label{tab:mvtec_3d_cross_class}
\end{table*}

\begin{table*}[!ht]
\centering
\fontsize{9}{10}\selectfont{
\begin{tabular}{c|ccc|c}
\toprule
Method$\rightarrow$      & CPR                 & HETMM               & Dinomaly            & \textbf{SNARM}              \\
Category$\downarrow$ & \cite{li2023target} & \cite{chen2024hard}&\cite{guo2024dinomaly}&($T=10^4$)\\\bottomrule
candle      &25.7/92.4/96.4/{\color{blue}{\textbf{94.1}}}&{\color{blue}{\textbf{29.2}}}/{\color{blue}{\textbf{93.0}}}/{\color{blue}{\textbf{97.9}}}/87.8&0.7/46.6/81.0/62.6&{\color{red}{\textbf{51.0}}}/{\color{red}{\textbf{94.1}}}/{\color{red}{\textbf{99.6}}}/{\color{red}{\textbf{98.9}}}\\
capsules    &0.3/8.0/47.0/60.7&{\color{blue}{\textbf{21.9}}}/{\color{blue}{\textbf{82.8}}}/{\color{blue}{\textbf{95.5}}}/{\color{blue}{\textbf{64.3}}}&0.8/21.8/71.6/62.8&{\color{red}{\textbf{63.6}}}/{\color{red}{\textbf{97.2}}}/{\color{red}{\textbf{99.2}}}/{\color{red}{\textbf{98.0}}}\\
cashew      &63.9/{\color{blue}{\textbf{88.0}}}/{\color{blue}{\textbf{96.3}}}/{\color{blue}{\textbf{92.7}}}&{\color{blue}{\textbf{65.4}}}/87.9/92.4/91.2&16.4/45.4/82.8/46.6&{\color{red}{\textbf{86.8}}}/{\color{red}{\textbf{92.6}}}/{\color{red}{\textbf{99.8}}}/{\color{red}{\textbf{99.1}}}\\
chewinggum  &{\color{blue}{\textbf{64.0}}}/57.8/{\color{blue}{\textbf{96.9}}}/{\color{blue}{\textbf{97.3}}}&35.4/{\color{blue}{\textbf{87.3}}}/90.7/92.1&37.6/59.7/95.1/85.1&{\color{red}{\textbf{75.1}}}/{\color{red}{\textbf{90.8}}}/{\color{red}{\textbf{99.5}}}/{\color{red}{\textbf{99.4}}}\\
fryum       &43.8/84.9/95.6/{\color{blue}{\textbf{91.6}}}&{\color{blue}{\textbf{49.7}}}/{\color{blue}{\textbf{86.8}}}/{\color{blue}{\textbf{95.8}}}/91.0&17.5/56.5/90.5/65.1&{\color{red}{\textbf{50.8}}}/{\color{red}{\textbf{92.6}}}/{\color{red}{\textbf{97.5}}}/{\color{red}{\textbf{99.2}}}\\
macaroni1   &4.4/{\color{blue}{\textbf{90.2}}}/{\color{blue}{\textbf{98.8}}}/83.6&{\color{blue}{\textbf{16.1}}}/86.2/95.7/{\color{blue}{\textbf{84.1}}}&0.7/63.5/90.8/51.6&{\color{red}{\textbf{27.6}}}/{\color{red}{\textbf{98.6}}}/{\color{red}{\textbf{99.8}}}/{\color{red}{\textbf{98.0}}}\\
macaroni2   &0.6/77.0/{\color{blue}{\textbf{94.5}}}/66.6&{\color{blue}{\textbf{5.5}}}/{\color{blue}{\textbf{86.0}}}/94.2/{\color{blue}{\textbf{77.9}}}&0.3/56.7/86.9/48.5&{\color{red}{\textbf{17.2}}}/{\color{red}{\textbf{98.3}}}/{\color{red}{\textbf{99.7}}}/{\color{red}{\textbf{92.9}}}\\
pcb1        &38.9/{\color{red}{\textbf{95.3}}}/99.0/94.0&{\color{blue}{\textbf{83.4}}}/{\color{blue}{\textbf{94.1}}}/{\color{blue}{\textbf{99.4}}}/{\color{blue}{\textbf{96.7}}}&1.8/16.8/82.3/44.2&{\color{red}{\textbf{84.3}}}/93.7/{\color{red}{\textbf{99.5}}}/{\color{red}{\textbf{98.0}}}\\
pcb2        &14.6/{\color{blue}{\textbf{90.5}}}/{\color{red}{\textbf{99.0}}}/{\color{blue}{\textbf{92.5}}}&{\color{blue}{\textbf{15.9}}}/89.3/97.9/90.7&5.0/56.8/88.6/56.1&{\color{red}{\textbf{41.4}}}/{\color{red}{\textbf{90.6}}}/{\color{blue}{\textbf{98.7}}}/{\color{red}{\textbf{95.9}}}\\
pcb3        &{\color{blue}{\textbf{29.3}}}/83.1/{\color{blue}{\textbf{98.6}}}/90.7&18.9/{\color{blue}{\textbf{89.5}}}/{\color{blue}{\textbf{98.6}}}/{\color{blue}{\textbf{93.0}}}&1.4/47.5/82.8/52.7&{\color{red}{\textbf{51.8}}}/{\color{red}{\textbf{94.0}}}/{\color{red}{\textbf{98.8}}}/{\color{red}{\textbf{98.2}}}\\
pcb4        &12.9/68.0/90.9/97.1&{\color{red}{\textbf{47.6}}}/{\color{red}{\textbf{90.7}}}/{\color{red}{\textbf{98.1}}}/{\color{red}{\textbf{99.7}}}&6.1/63.7/88.3/54.0&{\color{blue}{\textbf{44.9}}}/{\color{blue}{\textbf{87.1}}}/{\color{red}{\textbf{98.1}}}/{\color{blue}{\textbf{99.1}}}\\
pipe\_fryum &71.4/95.1/{\color{blue}{\textbf{99.3}}}/93.0&{\color{blue}{\textbf{71.8}}}/{\color{blue}{\textbf{95.3}}}/99.1/{\color{blue}{\textbf{98.9}}}&29.6/82.7/95.5/84.3&{\color{red}{\textbf{77.6}}}/{\color{red}{\textbf{96.5}}}/{\color{red}{\textbf{99.7}}}/{\color{red}{\textbf{99.7}}}\\ \midrule
Average     &30.8/77.5/92.7/87.8&{\color{blue}{\textbf{38.4}}}/{\color{blue}{\textbf{89.1}}}/{\color{blue}{\textbf{96.3}}}/{\color{blue}{\textbf{89.0}}}&9.8/51.5/86.3/59.5&{\color{red}{\textbf{56.0}}}/{\color{red}{\textbf{93.8}}}/{\color{red}{\textbf{99.1}}}/{\color{red}{\textbf{98.0}}}\\
\bottomrule
\end{tabular}}
\caption{Per-Class Results on the  VisA~\cite{zou_spot--difference_2022} Dataset under the Cross-class setting with P-AP/PRO/P-AUROC/I-AUROC metrics.}
\label{tab:visa_cross_class}
\end{table*}

\end{document}